\documentclass[mathfont=cm,dvipsnames]{article} 
\usepackage{fullpage}
\usepackage{url}
\usepackage{authblk}
\usepackage[round]{natbib} 
\usepackage{mathtools}
\usepackage{booktabs} 
\usepackage{tikz} 
\usepackage{algorithmic}
\usepackage{amssymb}
\usepackage{amsthm}
\usepackage[ruled,vlined]{algorithm2e}
\usepackage{subfigure}
\usepackage{xcolor}

\usepackage{amsfonts}
\usepackage{nicefrac}      
\usepackage{microtype}     
\newtheorem{theorem}{Theorem}
\newtheorem{proposition}{Proposition}
\newtheorem{definition}{Definition}
\newtheorem{assumption}{Assumption}
\newcommand{\rl}{reinforcement learning}

\newcommand{\Qtar}{Q_0}
\newcommand{\mc}{\mathcal}
\newcommand{\mb}{\mathbb}
\newcommand{\STAC}{${\tt STAC}$}
\newcommand{\stddpg}{${\tt STDDPG}$}
\newcommand{\sac}{${\tt SAC}$}
\newcommand{\ac}{${\tt AC}$}
\newcommand{\stsac}{${\tt STSAC}$}
\newcommand{\ddpg}{${\tt DDPG}$}
\newcommand{\AL}{${\tt AL}$}
\newcommand{\CL}{${\tt CL}$}
\newcommand{\act}{actor-critic}
\title{Stackelberg Actor-Critic: Game-Theoretic Reinforcement Learning Algorithms}

\author[1]{Liyuan Zheng}
\author[1]{Tanner Fiez}
\author[2]{Zane Alumbaugh}
\author[1]{Benjamin Chasnov}
\author[1]{Lillian J. Ratliff}
\affil[1]{University of Washington}
\affil[2]{University of California, Santa Cruz}
\date{}

\begin{document}
\maketitle

\begin{abstract}

	The hierarchical interaction between the actor and critic in {\act} based {\rl} algorithms naturally lends itself to a game-theoretic interpretation. We adopt this viewpoint and model the actor and critic interaction as a two-player general-sum game with a leader-follower structure known as a Stackelberg game. Given this abstraction, we propose a meta-framework for Stackelberg {\act} algorithms where the leader player follows the total derivative of its objective instead of the usual individual gradient.
	From a theoretical standpoint, we develop a policy gradient theorem for the refined update and provide a local convergence guarantee for the Stackelberg {\act} algorithms to a local Stackelberg equilibrium. From an empirical standpoint, we demonstrate via simple examples that the learning dynamics we study mitigate cycling and accelerate convergence compared to the usual gradient dynamics given cost structures induced by {\act} formulations. Finally, experiments on OpenAI gym environments show that Stackelberg {\act} algorithms always perform at least as well and often significantly outperform the standard {\act} algorithm counterparts.
	
\end{abstract}

\section{Introduction}

The algorithmic techniques for {\rl} can be classified into policy-based, value-based, and {\act} methods~\citep{sutton2018reinforcement}. Policy-based methods directly optimize a parameterized policy to maximize the expected return, while value-based methods estimate the expected return and then infer an optimal policy from the value-function by selecting the maximizing actions.
Actor-critic methods bridge policy-based and value-based methods by learning the parameterized policy (actor) and the value-function (critic) together. In particular, {\act} methods learn a critic that approximates the expected return of the actor while concurrently learning an actor to optimize the expected return based on the critic's estimation.

In this paper, we adopt a game-theoretic perspective of {\act} {\rl} algorithms. To provide some relevant background from game theory, recall that 
Stackelberg games are a class of games that describe interactions between a leader and a follower~\citep{bacsar1998dynamic}.
In a Stackelberg game, the leader is distinguished by the ability to act before the follower. As a result of this structure, the leader optimizes its objective accounting for the anticipated response of the follower, while the follower selects a best response to the leader's action to optimize its own objective. 
The interaction between the actor and critic in {\rl} has an intrinsic hierarchical structure reminiscent of a Stackelberg game, which motivates our work to contribute a novel game-theoretic modeling framework along with theoretical and empirical results.

\textbf{Modeling Contributions.}
We explicitly cast the interaction between the actor and critic as a two-player general-sum Stackelberg game toward solving {\rl} problems.
Notably, this perspective deviates from the majority of work on {\act} {\rl} algorithms, which implicitly neglect the interaction structure by independently optimizing the actor and critic objectives using individual gradient dynamics.
In order to solve the game iteratively in a manner that reflects the interaction structure, we study learning dynamics in which the player deemed the leader updates its parameters using the total derivative of its objective defined using the implicit function theorem and the player deemed the follower updates using the typical individual gradient dynamics. We refer to this gradient-based learning method as the Stackelberg gradient dynamics. The designations of leader and follower between the actor and critic can result in distinct game-theoretic outcomes and we explore both choices and explain how the proper roles depend on the respective objective functions.

\textbf{Theoretical Contributions.}
The Stackelberg gradient dynamics were previously studied in general nonconvex games and enjoy a number of theoretical guarantees~\citep{fiez2020implicit}. In this paper we tailor the analysis of this learning dynamic to the {\rl} problem.
To do this, we begin by developing a policy gradient theorem for the total derivative update (Theorem~\ref{THM:STACKPOLICYGRAD}). Then, building off of this result, we develop a meta-framework of Stackelberg {\act} algorithms. Specifically, this framework adapts the standard {\act}, deep deterministic policy gradient, and soft-actor critic algorithms to be optimized using the Stackelberg gradient dynamics in place of the usual individual gradient dynamics. 
For the Stackelberg {\act} algorithms this meta-framework admits, we prove local convergence (Theorem~\ref{THM:CONVERGENCE}) to local Stackelberg equilibrium.

\textbf{Experimental Contributions.}
From an empirical standpoint, we begin by pointing out in Section~\ref{sec:motivation} that the objective functions in {\act} algorithms commonly exhibit a type of hidden structure in terms of the parameters. Given this observation, we develop simple, yet illustrative examples comparing the behavior of Stackelberg {\act} algorithms with standard {\act} algorithms. In particular, we observe that the Stackelberg dynamics mitigate cycling in the parameter space and accelerate convergence. We discover from extensive experiments on OpenAI gym environments that similar observations carry over to complex problems and that our Stackelberg {\act} algorithms always perform at least as well and often significantly outperform the standard {\act} algorithm counterparts. 

\section{Related Work}
Game-theoretic frameworks have been studied extensively in {\rl} but mostly in multi-agent setting~\citep{yang2020overview}. In multi-agent {\rl}, the decentralized learning scheme is mostly adopted in practice~\citep{zhang2019multi}, where agents typically behave independently and optimize their own objective with no explicit information exchange. A shortcoming of this method is that agents fail to consider the learning process of other agents and simply treat them as a static component of the environment~\citep{hernandez2017survey}. To resolve this, several works design learning algorithms that explicitly account for the learning behavior of other agents~\citep{zhang2010multi, foerster2017learning,letcher2018stable}, which is shown to improve learning stability and induce cooperation. In contrast, \citet{prajapat2020competitive} study a competitive policy optimization method for multi-agent {\rl} which performs recursive reasoning about the behavior of opponents to exploit them in two-player zero-sum games. 
\citet{zhang2020bi} study multi-agent {\rl} problems, where each agent is using a typical actor-critic algorithm, with the twist that the follower’s policy takes the leader’s action as an input, which is used to approximate the potential best response. However, the procedure reduces to the usual actor-critic algorithm when applied to a single-agent {\rl} problem.

The past research taking a game-theoretic viewpoint of single-agent {\rl} is limited despite the fact that there is often implicitly multiple players in {\rl} algorithms. \citet{rajeswaran2020game} propose a framework that casts model-based {\rl} as a two-player general-sum Stackelberg game between a policy player and a model player. However, they only consider optimizing the objective of each player using the typical individual gradient dynamics with timescale separation as an approximation to Stackelberg gradient dynamics. Concurrent with this work, \citet{wen2021characterizing} show that Stackelberg policy gradient recovers the standard policy gradient under certain strong assumptions, including that the critic is directly parameterized by the $Q$-value function.
\citet{hong2020two} analyze the Stackelberg gradient dynamics with timescale separation for bilevel optimization with application to {\rl}. For {\rl}, they give a convergence guarantee for an actor-critic algorithm under assumptions such as exact linear function approximation which result in the total derivative being equivalent to the individual gradient. We provide a complimentary study by developing a general framework for Stackelberg {\act} algorithms that we analyze without such assumptions and also extensively evaluate empirically on {\rl} tasks.

\section{Motivation \& Preliminaries}
\label{sec:motivation}
In this section, we begin by presenting background on Stackelberg games and the relevant equilibrium concept.
Then, to motivate and illustrate the utility of Stackelberg-based {\act} algorithms, we highlight a key hidden structure that exists in actor-critic objective formulations and explore the behavior of Stackelberg gradient dynamics in comparison to individual gradient dynamics given this design.
Finally, we provide the necessary mathematical background and formalism for {\act}  {\rl} algorithms.

\subsection{Game-Theoretic Preliminaries}
\label{sec:stackgame}

A Stackelberg game is a game  between two agents where one agent is deemed the leader and the other the follower.  Each agent has an objective they want to optimize that depends on not only their own actions but also on the actions of the other agent. Specifically, the leader optimizes its objective under the assumption that the follower will play a best response. Let $f_1(x_1, x_2)$ and $f_2(x_1, x_2)$ be the objective functions that the leader and follower want to minimize, respectively, where $x_1\in X_1\subseteq\mb{R}^{d_1}$ and $x_2\in X_2\subseteq\mb{R}^{d_2}$ are their decision variables or strategies and $x=(x_1,x_2)\in X_1\times X_2$ is their joint strategy. 
The leader and follower aim to solve the following problems:
\begin{align}
	&\mkern-9mu\textstyle \min_{x_1\in X_1}\{f_1(x_1,x_2)\big|\ x_2\in \arg\min_{y\in
		X_2}f_2(x_1,y)\},\mkern-4mu \tag{ L}\\
	&\mkern-9mu\textstyle \min_{x_2\in X_2}f_2(x_1,x_2).\mkern-4mu\tag{ F}
\end{align}
Since the leader assumes the follower chooses a best response $x_2^*(x_1) = \arg\min_{y} f_2 (x_1, y)$,\footnote{Under sufficient regularity conditions on the follower's optimization problem, the best response map is a singleton.
	This is a generic condition in games~\citep{ratliff2014genericity,fiez2020implicit}.
} 
the follower’s decision variables are implicitly a function of the leader’s. In deriving sufficient conditions for the optimization problem in (L), the leader utilizes this information by the total derivative of its cost function which is given by
\begin{equation*}
	\nabla f_1(x_1, x_2^*(x_1)) = \nabla_{1}f_1(x) + (\nabla x_2^*(x_1))^\top\nabla_2f_1(x).
	\label{eq:stac_grad}
\end{equation*}
where
$\nabla x_2^*(x_1) = - ( \nabla_2^2f_2(x))^{-1}\nabla_{21}f_2(x)$.
\footnote{The partial derivative of $f(x_1,x_2)$ with respect to the $x_i$ is denoted by $\nabla_if(x_1,x_2)$ and the total derivative of $f(x_1,h(x_1))$ for some function $h$, is denoted $\nabla f$ where $\nabla f(x_1,h(x_1)=\nabla_1f(x_1,h(x_1))+(\nabla h(x_1))^\top \nabla_2f(x_1,h(x_1))$.}

	\begin{figure*}[t!]	
	\centering	
	
	\subfigure[][Individual gradient]{\label{fig:traj_gd}\includegraphics[width=0.25\linewidth]{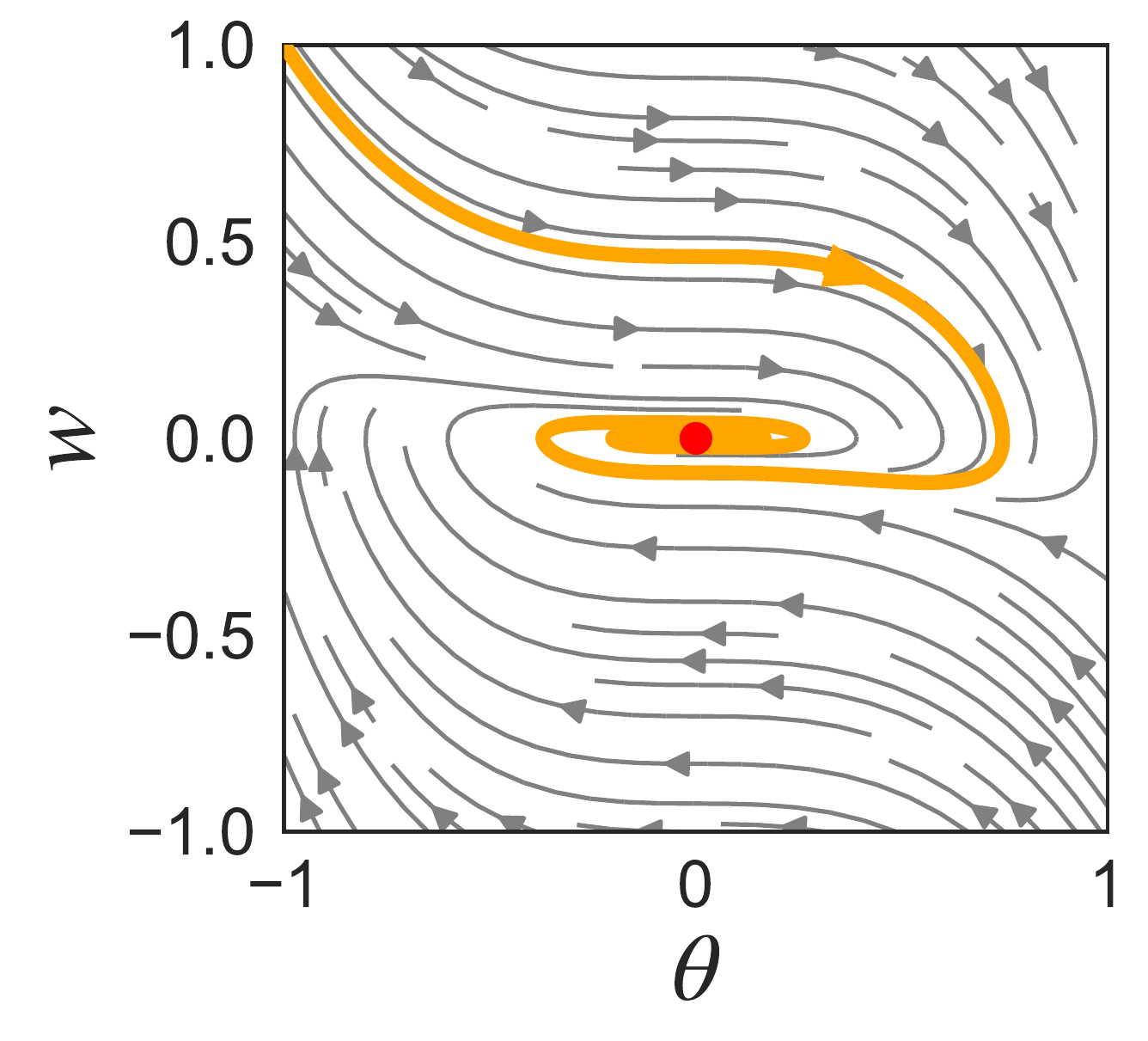}} \hfill	
	\subfigure[][Stackelberg gradient]{\label{fig:traj_stgd}\includegraphics[width=0.25\linewidth]{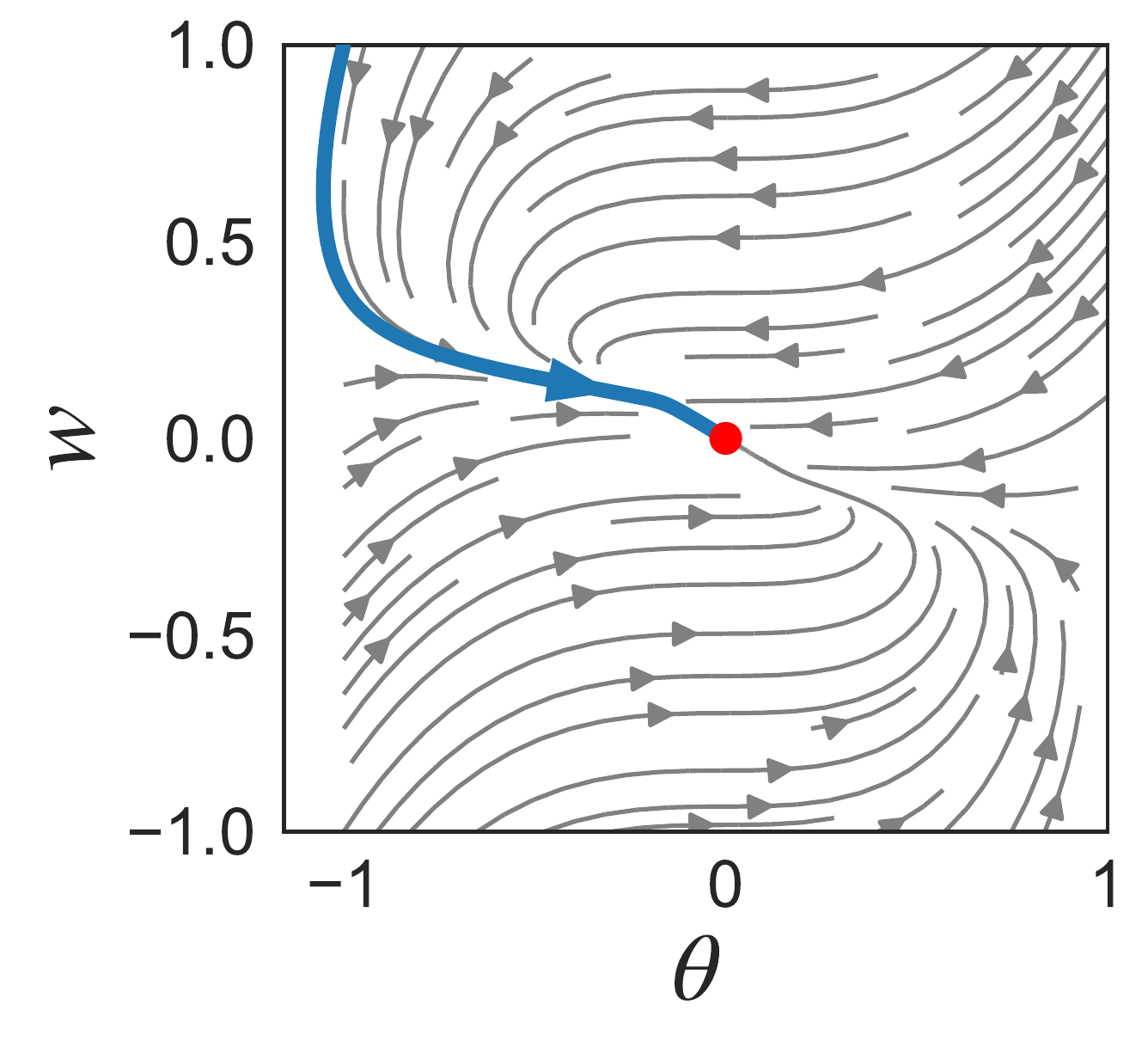}}\hfill	
	\subfigure[][Error to equilibrium]{\label{fig:error}\includegraphics[width=0.3\linewidth]{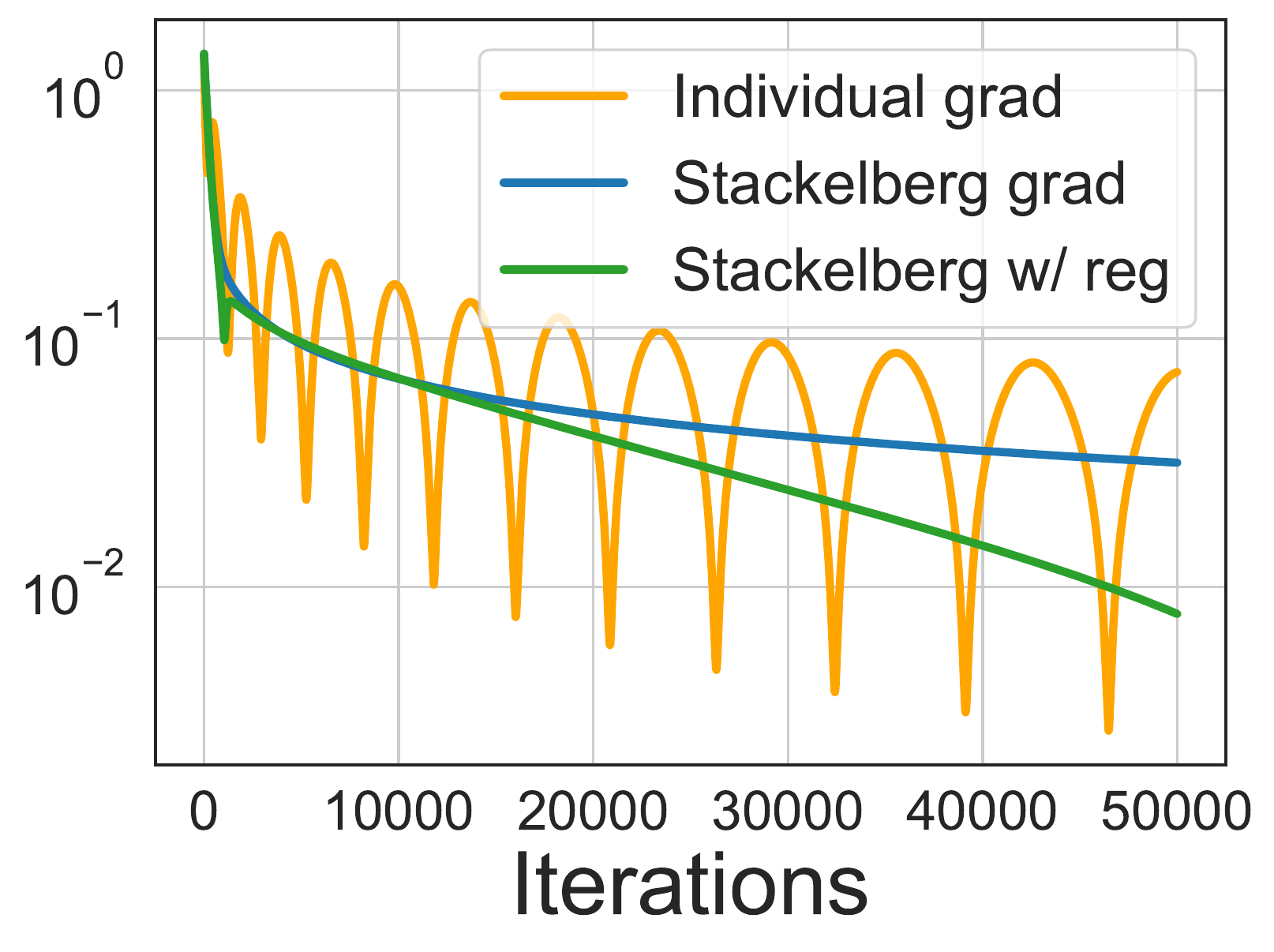}} 	
	\hfill	
	
	\caption{(a)--(b) Vector fields and trajectories of the actor and critic updates using individual gradient and Stackelberg gradient. (c) Error  $\|w-w^*\|^2 + \|\theta - \theta^*\|^2$ for individual gradient, Stackelberg gradient, and Stackelberg gradient with regularization, where $(\theta^\ast,w^\ast)=(0,0)$. }	
\end{figure*}	
\begin{figure*}[t!]	
	\centering	
	
	\subfigure[][Individual gradient]{\label{fig:traj_gd_sac}\includegraphics[width=0.25\linewidth]{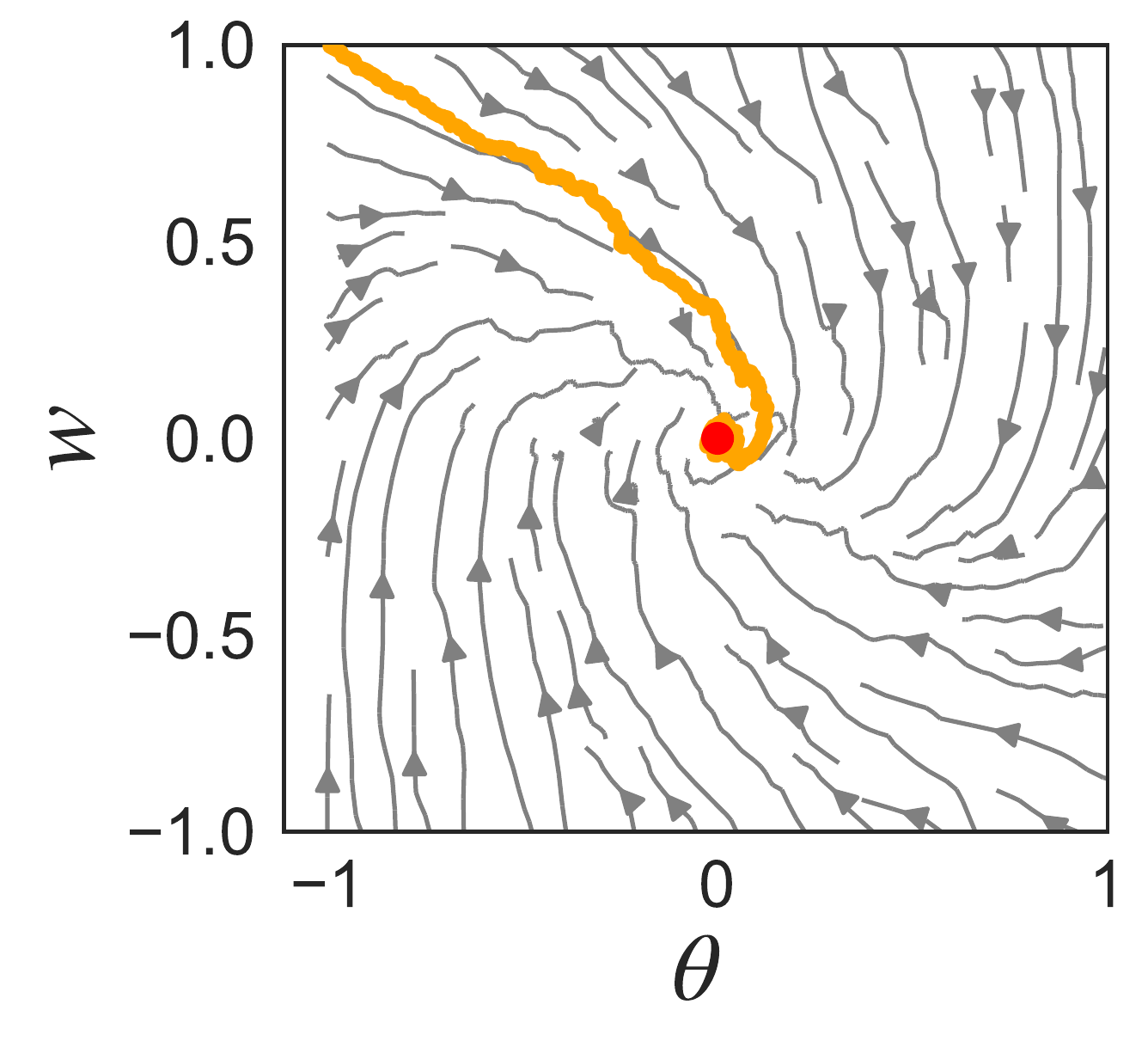}}\subfigure[][Stackelberg gradient]{\label{fig:traj_stgd_sac}\includegraphics[width=0.25\linewidth]{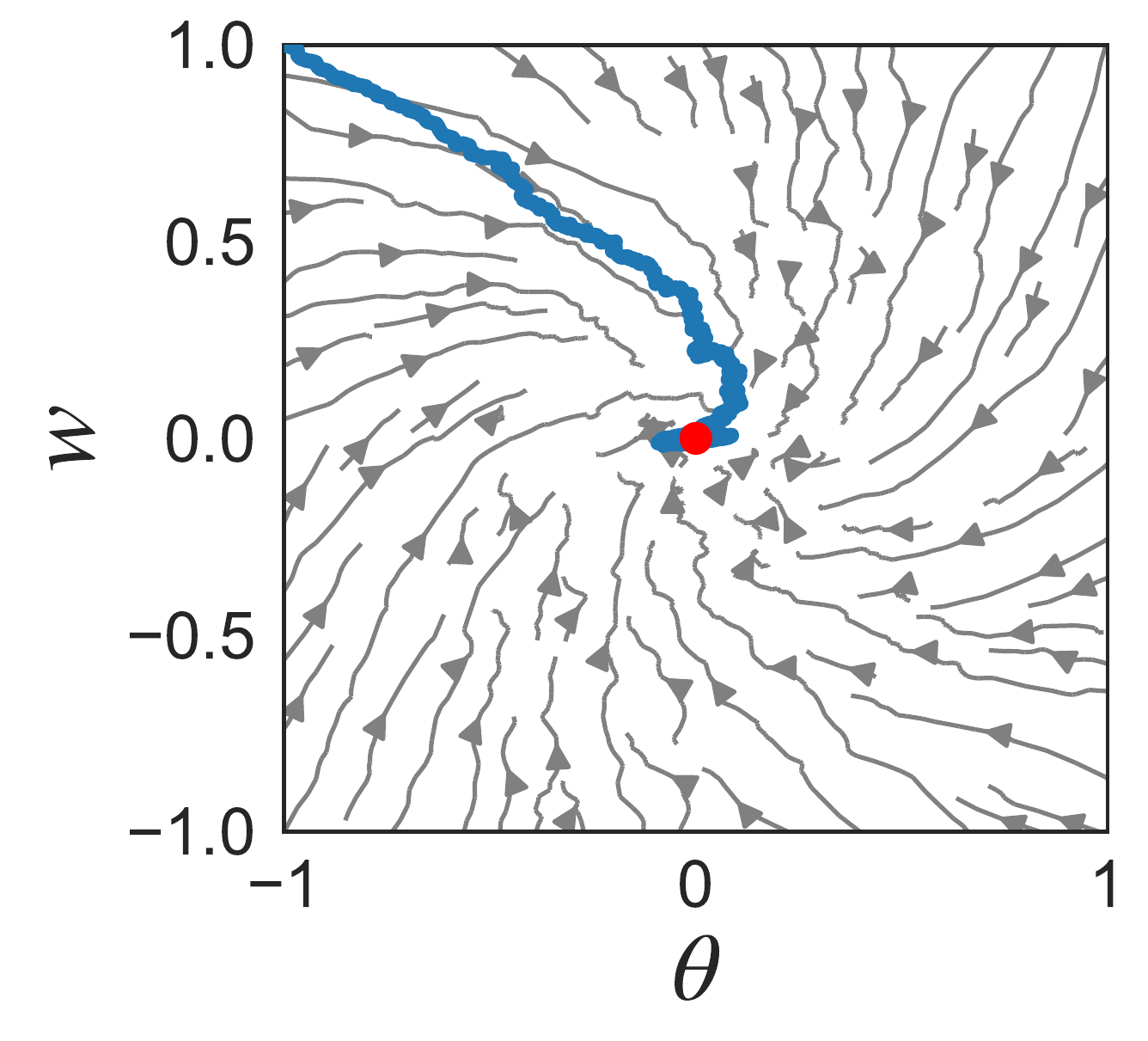}} 	
	\caption{(a)--(b) Individual gradient and Stackelberg gradient with entropic regularization in actor objective. }	
\end{figure*}

Hence, a point $x=(x_1,x_2)$ is a local solution to (L) if $\nabla f_1(x_1,x_2^\ast(x_1))=0$ and $\nabla^2 f_1(x_1,x_2^\ast(x_1))>0$. For the follower's problem, sufficient conditions for optimality are $\nabla_2f_2(x_1,x_2)=0$ and $\nabla_2^2f_2(x_1,x_2)>0$. This gives rise to the following equilibrium concept which characterizes sufficient conditions for a local Stackelberg equilibrium.
\begin{definition}[Differential Stackelberg Equilibrium, \citealt{fiez2020implicit}]
	The joint strategy $x^{\ast} = (x_1^\ast,x_2^\ast)\in X_1\times X_2$ is a differential Stackelberg equilibrium if $\nabla f_1(x^\ast)=0$, $\nabla_2f_2(x^\ast)=0$, $\nabla^2f_1(x^\ast)>0$, and $\nabla_2^2f_2(x^\ast)>0$.
	\label{def:stackelberg}
\end{definition}
The Stackelberg learning dynamics derive from the first-order gradient-based sufficient conditions and are given by
\begin{align*}
	x_{1,k+1}&=x_{1,k}-\alpha_1 \nabla f_1(x_{1,k},x_{2,k})\\
	x_{2,k+1}&=x_{2,k}-\alpha_2 \nabla_2 f_2(x_{1,k},x_{2,k})\
\end{align*}
where $\alpha_i$, $i=1,2$ are the leader and follower learning rates.

\iffalse\begin{definition}[Local Stackelberg ({\lse}) \citep{fiez2020implicit}]
	Consider $U_i\subset X_i$ for each $i\in \{1,2\}$. The strategy $x_1^\ast\in U_1$ is a local Stackelberg solution for the leader  if, $\forall x_1\in U_1$,
	\[\textstyle
	\sup_{x_2\in \reac_{U_2}(x_1^\ast)}  f_1(x_1^\ast, x_2)\leq \sup_{x_2\in
		\reac_{U_2}(x_1)}f_1(x_1,x_2),
	\]
	where $\reac_{U_2}(x_1)=\{y\in U_2|f_2(x_1,y)\leq f_2(x_1,x_{2}),  \forall   x_2\in U_2\}$.
	Moreover, $(x_1^\ast, x_2^\ast)$ for any $x_2^\ast\in \reac_{U_2}(x_1^\ast)$ is a local Stackelberg equilibrium on $U_1\times U_2$.
	\label{def:lse}
\end{definition}
\fi

\subsection{Motivating Examples} \label{sec:example}

In the next section we present several common actor-critic formulations including the ``vanilla'' {\act}, deep deterministic policy gradient, and soft {\act}. 
A common theme among them is that the actor and critic objectives exhibit a simple hidden structure in the parameters.
In particular, the actor objective typically has a hidden linear structure in terms of the parameters $\theta$ which is abstractly of the form $Q_w(\theta)=w^\top\mu(\theta)$. Analogously, the critic objective usually has a hidden quadratic structure in the parameters $w$ which is abstractly of the form or $(R(\theta)-Q_w(\theta))^2$. The terminology of hidden structure in this context refers to the fact that the specified structure appears when the functions transforming the parameters are removed.\footnote{The actor and critic functions could be approximated by neural nets in practice but we consider the simplest linear case, which captures the hidden structure and gives insights for general cases.} Interestingly, similar observations have been made regarding generative adversarial network formulations and exploited to gain insights into gradient learning dynamics for optimizing them~\citep{flokas2019poincar, flokas2021solving}.

Based on this observation, we investigate simple, yet illustrative {\rl} problems with the aforementioned structure and compare and contrast the behavior of the Stackelberg gradient dynamics with the usual individual gradient dynamics. As we demonstrate later in Section~\ref{sec:exp}, the insights we uncover from this study generally carry over to complex {\rl} problems.

\paragraph{Example.} Consider a single step Markov decision process where the reward function is given by $R(\theta) = -\frac{1}{5}\theta^2$ and $\theta \in [-1, 1]$ is the decision variable of actor. Suppose that the critic is designed using the most basic linear function approximation $Q_w(\theta) = w  \theta$ with $w \in [-1, 1]$.  
The actor seeks to find the action that maximizes the value indicated by the critic and the critic approximates the rewards of actions generated by the actor. Thus, the actor has objective $J(\theta, w) = Q_w(\theta) = w \theta$ and the critic has objective  $L(\theta, w) = \mathbb{E}_{\theta \sim \rho} [(R(\theta) - Q_w(\theta))^2]$. For simplicity, we assume the critic only minimizes the mean square error of the sample action generated by current actor $\theta$. The critic objective  is then $L(\theta, w) = (R(\theta) - Q_w(\theta))^2 = (w \cdot \theta + \frac{1}{5} \theta^2)^2$. 

\paragraph{Actor-Critic \& Deep Deterministic Policy Gradient.}
The structure of this example closely mirrors the hidden structure of both the ``vanilla'' actor-critic and deep deterministic policy gradient formulations as described in the next section. The typical way to optimize the objectives is by performing individual gradient dynamics (gradient descent on each cost) on the actor and critic parameters. Figure~\ref{fig:traj_gd} shows the gradient vector field and the parameter trajectories under the individual gradient dynamics. We observe that although the trajectory eventually converges to the equilibrium point $(\theta^*, w^*) = (0, 0)$, it cycles significantly.
Figure~\ref{fig:traj_stgd} shows the vector field and parameter trajectories under the Stackelberg gradient dynamics, the details of which will be introduced in Section~\ref{sec:stac_frame}. We observe that the cycling behavior is completely eliminated as a result of the consideration given to the interaction structure. 
Figure~\ref{fig:error} shows the error to equilibrium $\|w-w^*\|^2 + \|\theta - \theta^*\|^2$ for the individual gradient dynamics and the Stackelberg gradient dynamics along with a regularized version introduced in Section~\ref{sec:regularization}. This highlights that cycling is mitigated and convergence accelerated by optimizing using the Stackelberg gradient.

\paragraph{Soft Actor-Critic.}
The soft actor-critic algorithm also exhibits a similar structure, but with entropic regularization included in the actor objective. We show the vector fields along with the parameter trajectories for the individual gradient dynamics and the Stackelberg gradient dynamics in Figure~\ref{fig:traj_gd_sac} and Figure~\ref{fig:traj_stgd_sac}, respectively. Given the entropic regularization, both learning algorithms behave similarly. This perhaps indicates that the individual gradient dynamics are more well-suited to optimize this form of objectives and highlights the importance of considering how game dynamics perform on types of hidden structures when optimizing actor-critic algorithms in {\rl}.

Further details on the examples in this section are provided in Appendix~\ref{appendix:example}. Importantly, regardless of the objective function structure, the Stackelberg gradient dynamics tend to converge rather directly to the equilibrium and for some hidden structures they significantly mitigate oscillations and stabilize training. It is well-known that this is a desirable property of the {\rl} algorithms owing to the implications for both evaluation and real-world applications~\citep{chan2019measuring}. 
Together, this motivating section suggests that introducing the Stackelberg dynamics as a ``meta-algorithm'' on existing {\act} methods is likely to lead to more favorable convergence properties. We demonstrate this empirically in Section~\ref{sec:exp}, while now we introduce actor-critic algorithms.

\subsection{Actor-Critic Algorithms} \label{sec:ac_background}

We consider discrete-time Markov decision processes (MDPs) with continuous state space $\mathcal{S}$ and continuous action space $\mathcal{A}$. We denote the state and action at time step $t$ by $s_t$ and $a_t$, respectively. The initial state $s_0$ is determined by the initial state density $s_0 \sim \rho(s)$. At time step $t$, the agent in state $s_t$ takes an action $a_t$ according to a policy $a_t \sim \pi(\cdot|s_t)$ and obtains a reward $r_t = r(s_t, a_t)$. The agent then transitions to state $s_{t+1}$ determined by the transition function $s_{t+1} \sim P(s' |s_t, a_t)$. A trajectory $\tau = (s_0, a_0, \dots, s_T, a_T)$ gives the cumulative rewards or return defined as $R(\tau) =  \sum_{t=0}^T \gamma^t r(s_t, a_t)$, where the discount factor $0 < \gamma \le 1$ assigns weights to rewards received at different time steps. The expected return of $\pi$ after executing $a_t$ in state $s_t$ can be expressed by the $Q$ function 
\begin{equation}
	\textstyle	Q^\pi(s_t, a_t) = \mathbb{E}_{\tau \sim \pi} \big[ \sum_{t' = t}^T \gamma^{t'-t} r(s_{t'}, a_{t'}) | s_t, a_t \big].\notag 
\end{equation}
Correspondingly, the expected return of $\pi$ in state $s_t$ can be expressed by the value function $V$ defined as
\begin{equation}
	\textstyle	V^\pi(s_t) = \mathbb{E}_{\tau \sim \pi} \big[ \sum_{t' = t}^T \gamma^{t'-t} r(s_{t'}, a_{t'}) | s_t \big]. \notag 
\end{equation}
The goal of {\rl} is to find an optimal policy that maximizes the expected return which is given by
\begin{align}
	J(\pi)  &= \textstyle \mathbb{E}_{\tau \sim \pi}  \big[ \sum_{t = 0}^T \gamma^{t} r(s_t, a_t) \big] = \int_{\tau} p(\tau | \pi) R(\tau) \mathrm{d} \tau \nonumber \\
	& = \textstyle \mathbb{E}_{s \sim \rho, a \sim \pi(\cdot|s)} \big[ Q^\pi(s, a) \big],\notag
\end{align}
where $p(\tau | \pi) = \rho(s_0) \prod_{t=0}^T \pi(a_t|s_t) P(s_{t+1}|s_t, a_t)$.

The policy-based approach \citep{williams1992simple} parameterizes the policy $\pi$ by the parameter $\theta$ and finds the optimal parameter choice $\theta^\ast$ by maximizing the expected return 
\begin{equation}
	\textstyle J(\theta) = \mathbb{E}_{s \sim \rho, a \sim \pi_\theta(\cdot|s)} \big[ Q^\pi(s, a) \big]. \label{eq:rl_obj}
\end{equation}
This optimization problem can be solved by gradient ascent.
By the policy gradient theorem~\citep{sutton2000policy},
\begin{equation*}
	\nabla_\theta J(\theta) = \mathbb{E}_{s \sim \rho, a \sim \pi_\theta(\cdot|s)} \left[ \nabla_\theta \log \pi_\theta (a|s) Q^\pi(s, a) \right],
\end{equation*}
where $\nabla_\theta$ denotes the derivative with respect to $\theta$. A common method to approximate $Q^\pi(s, a)$ in the policy gradient is by sampling trajectories and averaging returns, which is known as REINFORCE \citep{williams1992simple}.

\paragraph{``Vanilla'' Actor-Critic ({\ac}).}
The {\act} method \citep{konda2000actor, grondman2012survey} relies on a critic function $Q_w(s, a)$ parameterized by $w$ to approximate $Q^\pi(s, a)$. By replacing $Q_w(s, a)$ with $Q^\pi(s, a)$ in~\eqref{eq:rl_obj}, the actor which is parameterized by $\theta$ has the objective
\begin{equation}
	\textstyle	 J(\theta,w) = \mathbb{E}_{s \sim \rho, a \sim \pi_\theta(\cdot|s)} \big[ Q_w(s, a) \big].
	\label{eq:actor_obj}
\end{equation}
The objective is optimized using gradient ascent where
\begin{equation}
	\mkern-13mu	\nabla_\theta J(\theta,w) = \mathbb{E}_{s \sim \rho, a \sim \pi_\theta(\cdot|s)} [ \nabla_\theta \log \pi_\theta (a|s) Q_w(s, a) ]. \label{eq:policy_grad}
\end{equation}
The critic which is parameterized by $w$ has the objective to minimize the mean square error between the $Q$-functions
\begin{equation}
	\textstyle	 L(\theta,w) = \textstyle \mathbb{E}_{s \sim \rho, a \sim \pi_\theta (\cdot | s)} [ (Q_w(s, a) - Q^\pi(s, a) )^2],
	\label{eq:critic_obj}
\end{equation}
where the function $Q^\pi(s, a)$ is approximated by Monte Carlo estimation or bootstrapping~\citep{sutton2018reinforcement}. 

The actor-critic method optimizes the objectives with individual gradient dynamics~\citep{peters2008natural, mnih2016asynchronous} which gives rise to the updates
\begin{align}
	\theta & \leftarrow \theta + \alpha_\theta \nabla_\theta J(\theta,w), \label{eq:grad_update1} \\
	w & \leftarrow w - \alpha_w \nabla_w L(\theta,w), \label{eq:grad_update2}
\end{align}
where $\alpha_\theta$ and $\alpha_w$ are the learning rates of actor and critic.
Clearly, even in this basic {\act} method, the actor and critic are coupled since $J$ and $L$ depend on both $\theta$ and $w$, which naturally lends to a game-theoretic interpretation. 

\paragraph{Deep Deterministic Policy Gradient ({\ddpg}).}
The {\ddpg} algorithm~\citep{lillicrap2015continuous} is an off-policy method with subtly different objective functions for the actor and critic. In particular, the formulation has a deterministic actor $\mu_\theta(s): \mathcal{S} \rightarrow \mathcal{A}$ with the objective
\begin{equation}
	J(\theta, w) = \mathbb{E}_{\xi \sim \mathcal{D}} \left[ Q_w(s, \mu_\theta(s)) \right].
	\label{eq:ddpg_actor}
\end{equation}
The critic objective is the mean square Bellman error
\begin{equation}
	\mkern-11mu L(\theta, w) =  \underset{\xi \sim \mathcal{D}}{\mathbb{E}} [ \left(Q_w(s, a) - (r + \gamma \Qtar(s', \mu_\theta(s'))) \right)^2 ],
	\label{eq:ddpg_critic}
\end{equation}
where $\xi=(s,a,r,s')$, $\mathcal{D}$ is a replay buffer, and $\Qtar$ is a target $Q$ network.\footnote{In the {\ddpg} algorithm, the next-state actions used in the target network come from the target policy instead of the current policy. To be consistent with {\sac}, we use the current policy.}

\paragraph{Soft Actor-Critic ({\sac}).}
The {\sac} algorithm~\citep{haarnoja2018soft} exploits the double Q-learning trick~\citep{van2016deep} and employs entropic regularization to encourage exploration. The actor's objective $J(\theta, w)$ is
\begin{equation}
	\begin{split}
		\mathbb{E}_{\xi \sim \mathcal{D}} \big[ \min_{i=1,2} Q_{w_i}(s, a_\theta(s)) -\eta \log(\pi_\theta(a_\theta(s)|s)) \big],  
	\end{split}
	\label{eq:sac_actor}
\end{equation}
where $a_\theta (s)$ is a sample from $\pi_\theta(\cdot|s)$ and $\eta$ is entropy regularization coefficient. The parameter of the critic is the union of both Q networks parameters $w = \{w_1, w_2\}$ and the critic objective is defined correspondingly by
\begin{equation}
	\mkern-4mu L(\theta, w) =  \textstyle \mathbb{E}_{\xi \sim \mathcal{D}} \big[ \sum_{i=1, 2} \left( Q_{w_i}(s, a) - y(r, s') \right)^2 \big],
	\label{eq:sac_critic}
\end{equation}
where
\[y(r, s')\mkern-2mu = \mkern-2mu r + \gamma (\min_{i=1,2} \mkern-3mu Q_{0,i}(s', a_\theta(s'))- \eta \log(\pi_\theta(a_\theta(s')|s'))).\]
The target networks in {\ddpg} and {\sac} are updated by taking the Polyak average of the network parameters over the course of training, and the actor and critic networks are updated by individual gradient dynamics identical to~\eqref{eq:grad_update1}--\eqref{eq:grad_update2}.

\section{Stackelberg Framework} \label{sec:stac_frame}
In this section, we begin by formulating the {\act} interaction as two-player general-sum Stackelberg game and introduce a Stackelberg framework for {\act} algorithms, under which we develop novel Stackelberg versions of existing algorithms: Stackelberg {\act} (\STAC), Stackelberg deep deterministic policy gradient (\stddpg), and Stackelberg soft {\act} (\stsac). Following this, we give a local convergence guarantee for the algorithms to a local Stackelberg equilibrium. Finally, a regularization method for practical usage of the algorithms is discussed.

\begin{algorithm}[t]
	\SetAlgoLined
	\KwIn{{\act} algorithm ${\tt ALG}$, player designations, and learning rate sequences $\alpha_{\theta,k},\alpha_{w,k}$.}
	\textbf{if} actor is leader, update actor and critic in ${\tt ALG}$ with:
	\vspace{-2mm}
	\begin{align}
		\theta_{k+1} & = \theta_k + \alpha_{\theta,k} \nabla J(\theta_k, w_k) \label{eq:stac_grad_update1}\\
		w_{k+1} & = w_k - \alpha_{w,k} \nabla_w L(\theta_k, w_k)\label{eq:stac_grad_update2}
	\end{align}
	\vspace{-7mm}
	
	\textbf{if} critic is leader, update actor and critic in ${\tt ALG}$ with:
	\vspace{-2mm}
	\begin{align}
		\theta_{k+1} & = \theta_k + \alpha_{\theta,k} \nabla_\theta J(\theta_k, w_k) \label{eq:switch1} \\
		w_{k+1} & = w_k - \alpha_{w,k} \nabla L(\theta_k, w_k)\label{eq:switch2}
	\end{align}
	\vspace{-5mm}
	\caption{Stackelberg Actor-Critic Framework
	}
	\label{alg:framework}
\end{algorithm}

\subsection{Meta-Algorithm}
\label{sec:meta}
Given an actor-critic formulation, in particular, the objectives of the actor and critic defined by $J(\theta, w)$ and $L(\theta, w)$, we can interpret the problem as a two-player general-sum Stackelberg game. If we view the actor as the leader and the critic as a follower, then the players aim to solve the following optimization problems, respectively:
\begin{align*}
	&\textstyle \max_{\theta}\{J(\theta, w^*(\theta)) \big| \ w^*(\theta) = \arg\min_{w'} L(\theta, w') \} \tag{AL} \\
	&\textstyle \min_{w} L(\theta, w). \tag{CF} 
\end{align*}
On the other hand, if we view the critic as the leader and the actor as the follower, then the players aim to solve the following optimization problems, respectively:
\begin{align*}
	&\mkern-9mu\textstyle\min_{w}\{L(\theta^{\ast}(w), w) \big| \ \theta^{\ast}(w) = \arg\max_{\theta'} J(\theta', w) \} \tag{CL} \\
	&\mkern-9mu\textstyle\max_{\theta} J(\theta, w). \tag{AF} 
\end{align*}

As described in Section~\ref{sec:stackgame}, we propose to optimize the objectives using a learning algorithm that accounts for the structure of the problems. Specifically, since the leader assumes the follower selects a best response, it is natural to optimize the leader objective by following the total derivative given that the follower's decision is implicitly a function of the leader's. 
The meta-framework we adopt for Stackelberg refinements of actor-critic methods is in Algorithm~\ref{alg:framework}. The distinction compared to the usual actor-critic methods is that in the updates we replace the individual gradient for the leader by the implicitly defined total derivative which accounts for the interaction structure whereas the rest of the actor-critic method remains identical.

The dynamics with the actor as the leader are given by~\eqref{eq:stac_grad_update1}--\eqref{eq:stac_grad_update2}
where the actor's total derivative $J(\theta,w)$ is 
\begin{equation}
	\mkern-10mu \nabla_\theta J(\theta, w) - \nabla_{w\theta}^\top L(\theta,w)(\nabla_w^2L(\theta,w))^{-1}\nabla_w J(\theta,w). \mkern-4mu \label{eq:stac_grad2}
\end{equation}
When the critic is the leader the dynamics are given by~\eqref{eq:switch1}--\eqref{eq:switch2}
where the critic's total derivative $\nabla L(\theta,w)$ is
\begin{equation}
	\mkern-10mu \nabla_w L(\theta,w)- \nabla_{\theta w}^\top J(\theta,w)(\nabla_\theta^2J(\theta,w))^{-1}\nabla_\theta L(\theta,w). \mkern-4mu\label{eq:stac_grad3}
\end{equation}

We now consider instantiations of this framework and explain how the total derivative can be obtained from sampling along with natural choices of leader and follower.

\subsection{Stackelberg ``Vanilla'' Actor-Critic}
We start by instantiating the Stackelberg meta-algorithm for the ``vanilla'' {\act} (\ac) algorithm for which the actor and critic objectives are given in~\eqref{eq:actor_obj} and~\eqref{eq:critic_obj}, respectively.\footnote{We only demonstrate the ``vanilla'' {\act} algorithm and its Stackelberg version here and in our experiments, but the framework could be generalized to more on-policy {\act} algorithms (e.g., A2C, A3C,~\citealt{mnih2016asynchronous}).} 
In this on-policy formulation, the critic assists the actor in learning the optimal policy by approximating the value function of the current policy. To give an accurate approximation, the critic aims to be selecting a best response $w^*(\theta) = \arg\min_{w'} L(\theta, w')$. Thus, the actor naturally plays the role of leader and the critic the follower. 

However, estimating the total derivative $\nabla J(\theta, w)$ as defined in~\eqref{eq:stac_grad2} is not straightforward and we analyze each component individually.
The individual gradient $\nabla_\theta J(\theta, w)$ can be computed by policy gradient theorem as given in~\eqref{eq:policy_grad}. Moreover, 
$\nabla_w J(\theta, w) = \mathbb{E}_{s \sim \rho, a \sim \pi_\theta (\cdot | s)} [\nabla_w Q_w(s, a)]$, which follows by direct computation,
and similarly 
\begin{align*}
	\nabla^2_w L(\theta, w)&=\mathbb{E}_{s \sim \rho, a \sim \pi_\theta (\cdot | s)}  \left[ 2\nabla_w Q_w(s, a)\nabla_w^\top Q_w(s, a) \right.\\
	& \quad\left. + 2(Q_w(s, a) - Q^\pi(s, a)) \nabla_w^2 Q_w(s,a) \right].
\end{align*}

To compute $\nabla_{w\theta} L(\theta, w)$ in \eqref{eq:stac_grad2}, we begin by obtaining $\nabla_\theta L(\theta, w)$ with the following policy gradient theorem. The proof of Theorem~\ref{THM:STACKPOLICYGRAD} is in Appendix~\ref{th1_proof}.
\begin{theorem} \label{THM:STACKPOLICYGRAD}
	Given an MDP and {\act} parameters $(\theta, w)$, the gradient of $L(\theta,w)$ with respect to $\theta$ is given by
	\begin{align*}
		& \textstyle	\nabla_\theta L(\theta, w) = \mathbb{E}_{\tau \sim \pi_\theta} [ \vphantom{\sum_{t=1}^T} \nabla_\theta \log \pi_\theta(a_0|s_0)  \nonumber \\
		& \textstyle	\quad (Q_w(s_0, a_0) - Q^\pi(s_0, a_0) )^2  + \sum_{t=1}^T \gamma^t \nabla_\theta \log \pi_\theta(a_t|s_t) \nonumber \\
		& \textstyle	\quad  (Q^\pi(s_0, a_0) - Q_w(s_0, a_0) ) Q^\pi(s_t,a_t) \vphantom{\sum_{t=1}^T} ].
	\end{align*}
\end{theorem}
Theorem~\ref{THM:STACKPOLICYGRAD} allows us to  compute $\nabla_{\theta w} L(\theta, w)$ directly by $\nabla_w(\nabla_\theta L(\theta,w))$ since the distribution of $\nabla_\theta L(\theta,w)$ does not depend on $w$ and $\nabla_w$ can be moved into the expectation. 

The critic in {\ac} is often designed to approximate the state value function $V^\pi(s)$ which has computational advantages, and the policy gradient can be computed by advantage estimation~\citep{schulman2015high}. In this formulation, $J(\theta,w) = \mathbb{E}_{\tau \sim \pi_\theta} \big[r(s_0, a_0) + V_w(s_1) \big]$ and $L(\theta, w) =  \mathbb{E}_{s \sim \rho} [ (V_w(s) - V^\pi(s) )^2 ]$. Then $\nabla_\theta L(\theta, w)$ can be computed by the next proposition that is derived in Appendix~\ref{prop1_proof}.
\begin{proposition} \label{PROP:STACKVALUEGRAD}
	Given an MDP and {\act} parameters $(\theta,w)$, if the critic has the objective function $L(\theta, w) =  \mathbb{E}_{s \sim \rho} [ (V_w(s) - V^\pi(s) )^2 ]$, then $\nabla_\theta L(\theta, w)$ is given by
	\begin{equation*}
		\underset{\tau \sim \pi_\theta}{\mathbb{E}} \mkern-8mu [ 2 \mkern-2mu \textstyle \sum\limits_{t=0}^T \mkern-4mu\gamma^t \nabla_\theta \log \pi_\theta(a_t|s_t) (V^\pi(s_0) - V_w(s_0) ) Q^\pi(s_t, a_t) \vphantom{\sum_{t=1}^T}].
	\end{equation*} 
\end{proposition}
Given these derivations, terms in~\eqref{eq:stac_grad2} can be estimated by sampled trajectories, and {\STAC} updates using~\eqref{eq:stac_grad_update1}--\eqref{eq:stac_grad_update2}.

\subsection{Stackelberg DDPG and SAC}
In comparison to on-policy methods where the critic is designed to evaluate the actor using sampled trajectories generated by the current policy, in off-policy methods the critic minimizes the Bellman error using samples from a replay buffer. Thus, the leader and follower designation between the actor and critic in off-policy methods is not as clear. To this end, we propose variants of {\stddpg} and {\stsac} where the leader and follower order can be switched. Given the actor as the leader (${\tt AL}$), the algorithms are similar to policy-based methods, where the critic plays an approximate best response to evaluate the current actor. On the other hand, given the critic as the leader (${\tt CL}$), the actor plays an approximate best response to the critic value, resulting in behavior closely resembling that of the value-based methods.

As shown in \eqref{eq:ddpg_actor}--\eqref{eq:ddpg_critic} for {\ddpg} and  \eqref{eq:sac_actor}--\eqref{eq:sac_critic} for {\sac}, the objective functions of off-policy methods are defined in expectation over an arbitrary distribution from a replay buffer instead of the distribution induced by the current policy. Thus, each terms in the total derivatives updates in~\eqref{eq:stac_grad2} and \eqref{eq:stac_grad3} can be computed directly and estimated by samples. Then, {\stddpg} and {\stsac} update using~\eqref{eq:stac_grad_update1}--\eqref{eq:stac_grad_update2} or~\eqref{eq:switch1}--\eqref{eq:switch2} depending on the choices of leader and follower.

\subsection{Convergence Guarantee} \label{sec:converge}
Consider, without loss of generality, the actor is designated as the leader and the critic the follower. Then, the actor and critic updates with the Stackelberg gradient dynamics and learning rates sequences $\{\alpha_{\theta, k}\},\{\alpha_{w, k}\}$ are of the form
\begin{align}
	\theta_{k+1} & = \theta_k + \alpha_{\theta, k} (\nabla J(\theta, w) + \epsilon_{\theta, k+1}), \label{eq:dtsys1}\\
	w_{k+1} & = w_k - \alpha_{w, k} (\nabla_w L(\theta, w) + \epsilon_{w, k+1}),\label{eq:dtsys2}
\end{align}
where $\{ \epsilon_{\theta, k+1}\}, \{\epsilon_{w, k+1}\}$ are stochastic processes.
The results in this section assume the following. 
\begin{assumption}
	The maps $\nabla J:\mb{R}^m \rightarrow \mb{R}^{m_\theta}$, $\nabla_w L:\mb{R}^m \rightarrow\mb{R}^{m_w}$ are Lipschitz, and $\|\nabla J\|<\infty$. The learning rate sequences are such that $\alpha_{\theta, k}=o(\alpha_{w, k})$ and $\sum_{k}\alpha_{i,k}=\infty$, $\sum_{k}\alpha_{i,k}^2<\infty$ for $i \in \mc{I}=\{ \theta, w\}$. The noise processes $\{\epsilon_{i,k}\}$ are zero mean, martingale difference sequences: given the filtration $\mc{F}_k=\sigma(\theta_s, w_s, \epsilon_{\theta,s}, \epsilon_{w,s},\ s\leq k)$, $\{\epsilon_{i,k}\}_{i\in \mc{I}}$ are conditionally independent, $\mb{E}[\epsilon_{i,k+1}|\ \mc{F}_k]=0$ a.s., and $\mb{E}[\|\epsilon_{i,k+1}\||\
	\mc{F}_{k}]\leq c_i(1+\|(\theta_{k}, w_k)\|)$ a.s.~for some constants
	$c_i\geq 0$ and  $i\in \mc{I}$. 
	\label{assump:convergence}
\end{assumption}
The following result gives a local convergence guarantee to a local Stackelberg equilibrium under the assumptions and the proof is in~Appendix~\ref{app_sec:convergenceproof}. For this result, recall that for a continuous-time dynamical system of the form $\dot{z}=-g(z)$, a stationary point $z^{\ast}$ of the system is said to be locally asymptotically stable or simply stable if the spectrum of the Jacobian denoted by $-Dg(z)$ is in the open left half plane. 
\begin{theorem}
	Consider an MDP and {\act} parameters $(\theta,w)$. 
	Given a locally asymptotically stable differential Stackleberg equilibrium $(\theta^\ast,w^\ast)$ of the continuous-time limiting system $(\dot{\theta},\dot{w})=(\nabla J(\theta,w),-\nabla_w L(\theta,w))$, under Assumption~\ref{assump:convergence} there exists a neighborhood $U$ for which the iterates $(\theta_k,w_k)$ of the discrete-time system in~\eqref{eq:dtsys1}--\eqref{eq:dtsys2} converge asymptotically almost surely to $(\theta^\ast,w^\ast)$ for $(\theta_0,w_0) \in U$. 
	\label{THM:CONVERGENCE}
\end{theorem}
This result is effectively giving the guarantee that the discrete-time dynamics locally converge to a stable, game theoretically meaningful equilibrium of the continuous-time system  using stochastic approximation methods given proper learning rates and unbiased gradient estimates~\citep{borkar2009stochastic}. 

\subsection{Implicit Map Regularization} \label{sec:regularization}
The total derivative in the Stackelberg gradient dynamics requires computing
the inverse of follower Hessian $\nabla^2_{2} f_2(x)$. 
Since critic networks in practical {\rl} problems may be highly non-convex, $(\nabla^2_{2} f_2(x))^{-1}$ can be ill-conditioned. 
Thus, instead of computing this term directly in the Stackelberg {\act} algorithms, we compute a regularized variant of the form $(\nabla^2_{2} f_2(x) + \lambda I)^{-1}$. This regularization method can be interpreted as the leader viewing the follower as optimizing a regularized cost $f_2(x)+\tfrac{\lambda}{2}\|x_2\|^2$,  while the follower actually optimizes $f_2(x)$.
The regularization $\lambda$ can interpolate between the Stackelberg and individual gradient updates for the leader as we now formalize.
\begin{proposition}
	Consider a Stackelberg game where the leader updates using the regularized total derivative $\nabla^{\lambda} f_1(x) = \nabla_{1}f_1(x) - \nabla_{21}^{\top}f_2(x) (\nabla_2^2f_2(x) + \lambda I)^{-1} \nabla_2f_1(x)$. As $\lambda \rightarrow 0$ then $\nabla^{\lambda}f_1(x)\rightarrow \nabla f_1(x)$ and when $\lambda \rightarrow \infty$ then $\nabla^{\lambda}f_1(x)\rightarrow \nabla_1 f_1(x)$.
\end{proposition}

\begin{figure*}[t!]
	\centering
	\subfigure{\label{fig:legend}\includegraphics[width=0.5\linewidth]{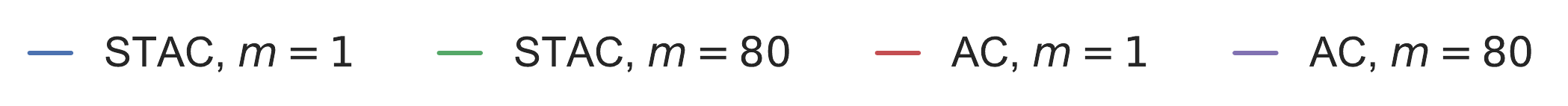}} \\
	\vspace{-4mm}
	\addtocounter{subfigure}{-1}
	\subfigure[${\tt CartPole}$]{\label{fig:cartpole1}\includegraphics[width=0.22\linewidth]{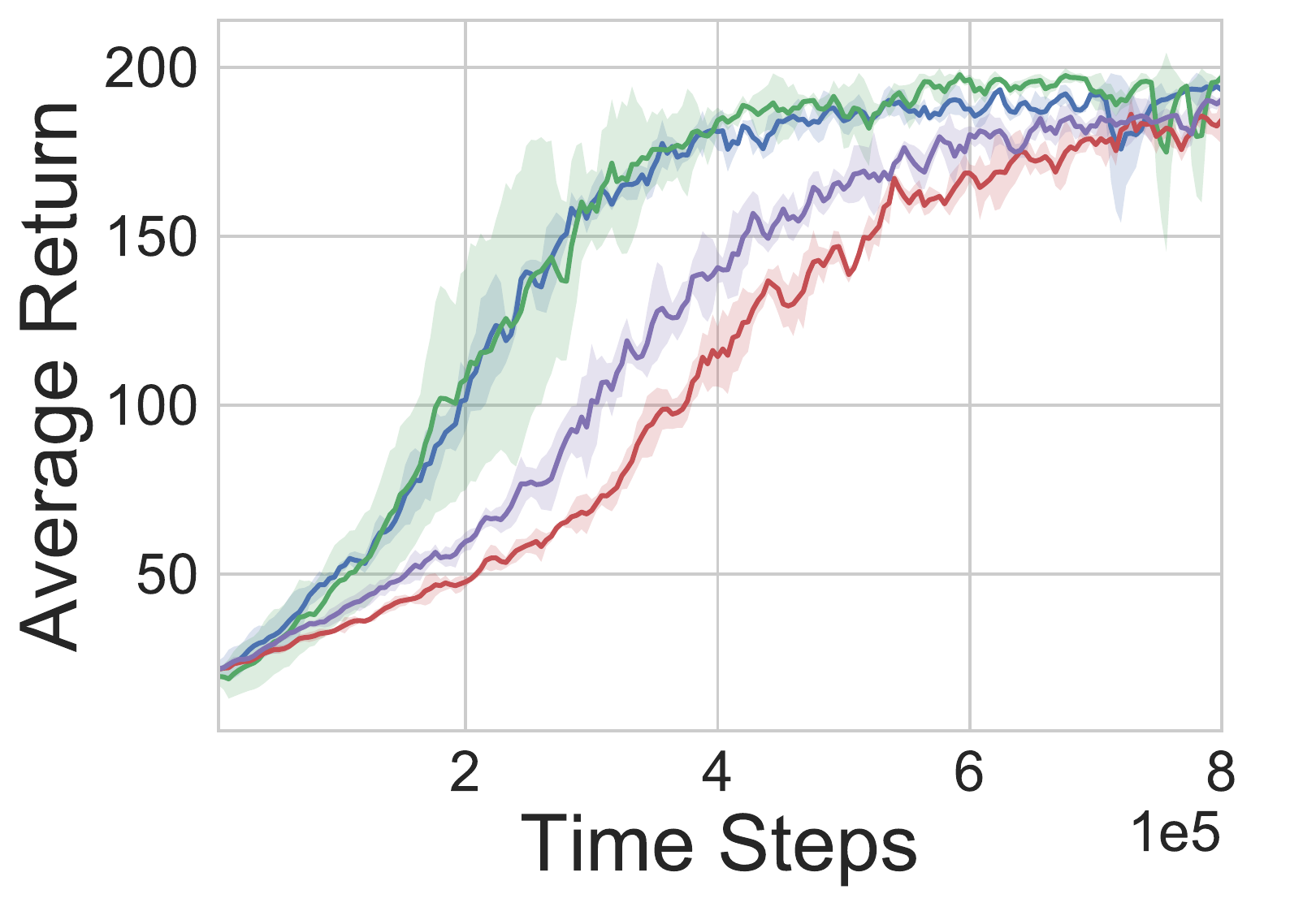}} \hfill
	\subfigure[${\tt Reacher}$]{\label{fig:reacher}\includegraphics[width=0.22\linewidth]{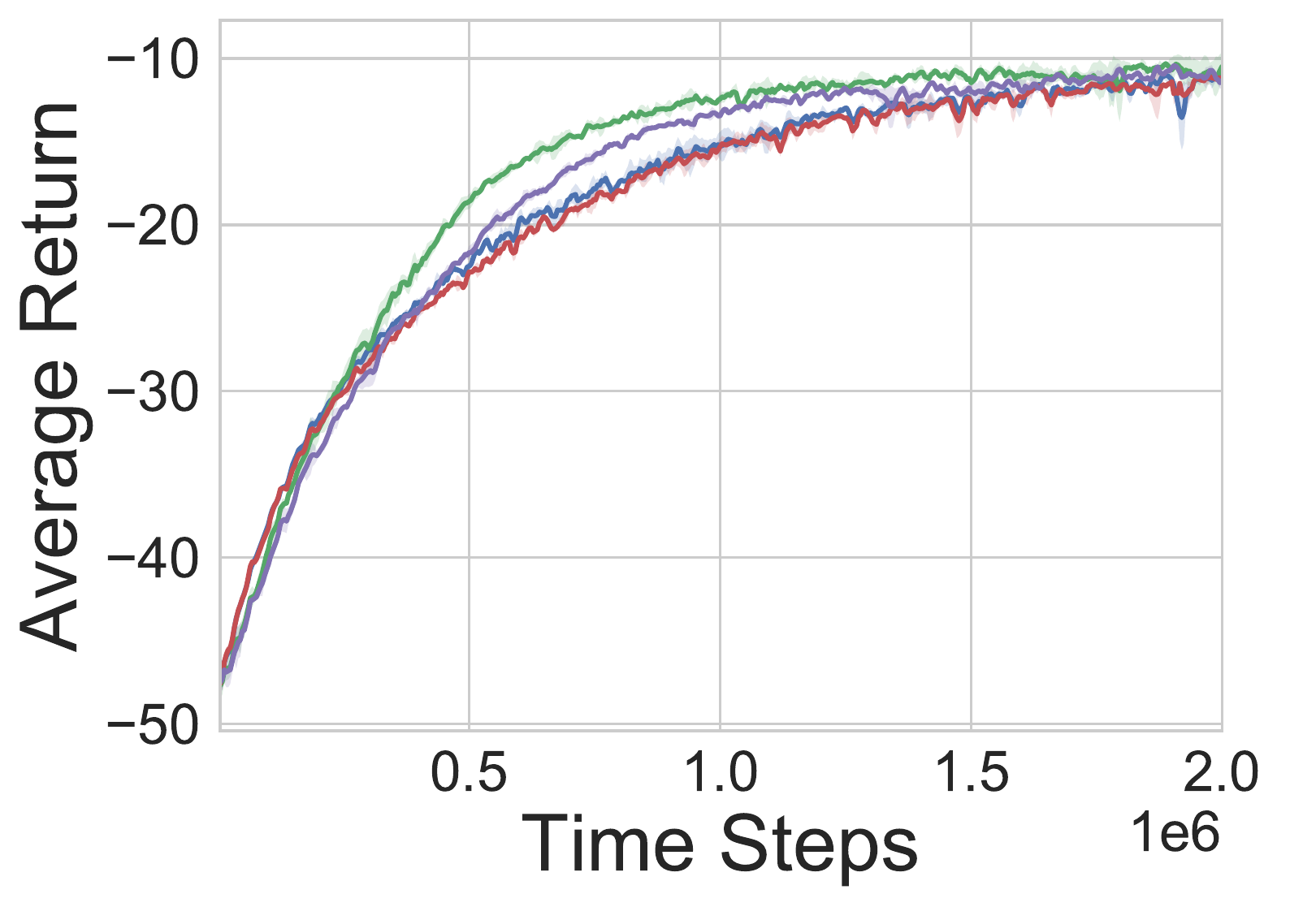}} \hfill
	\subfigure[${\tt Hopper}$]{\label{fig:hopper}\includegraphics[width=0.22\linewidth]{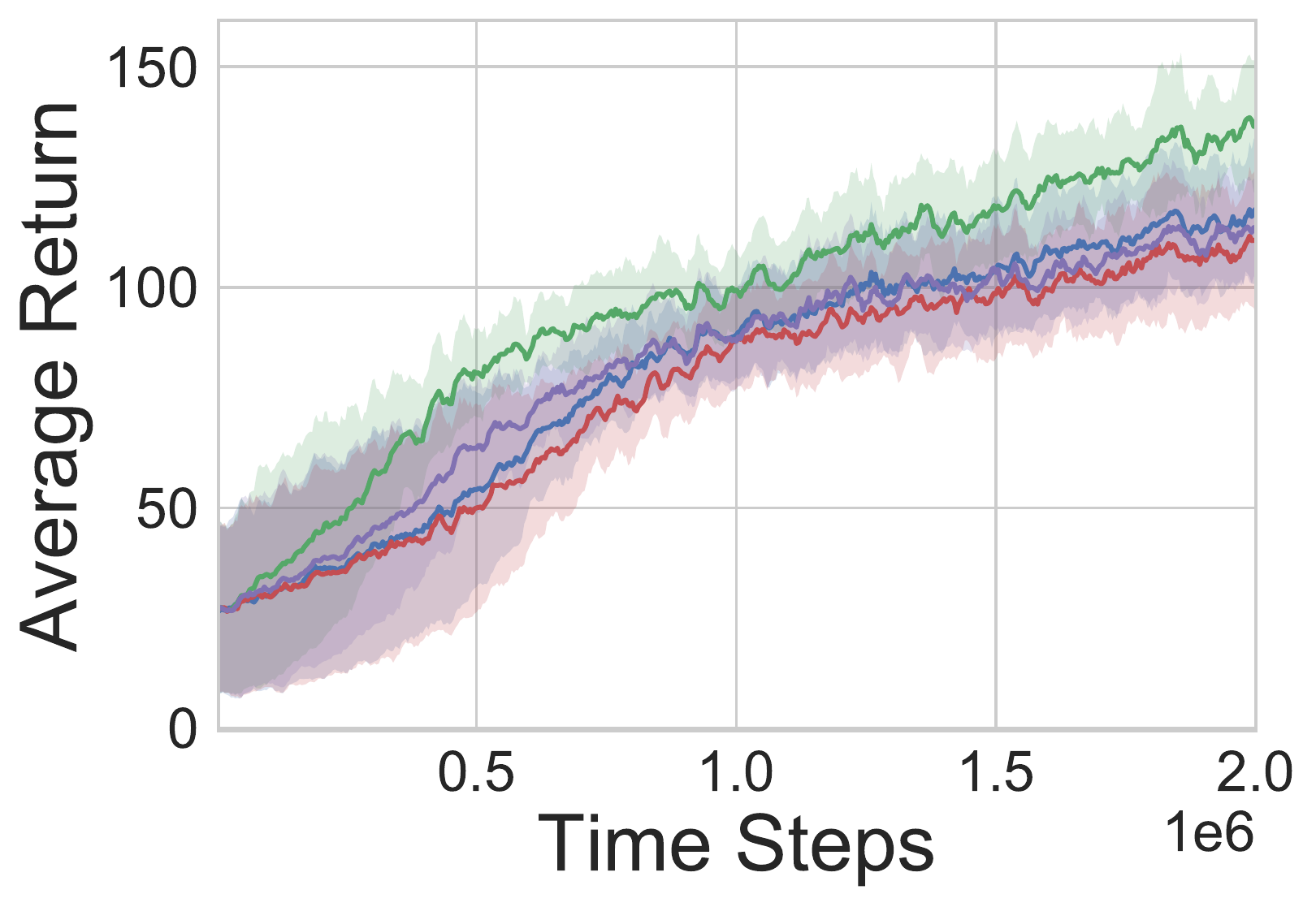}} \hfill
	\subfigure[${\tt Walker2d}$]{\label{fig:walker}\includegraphics[width=0.22\linewidth]{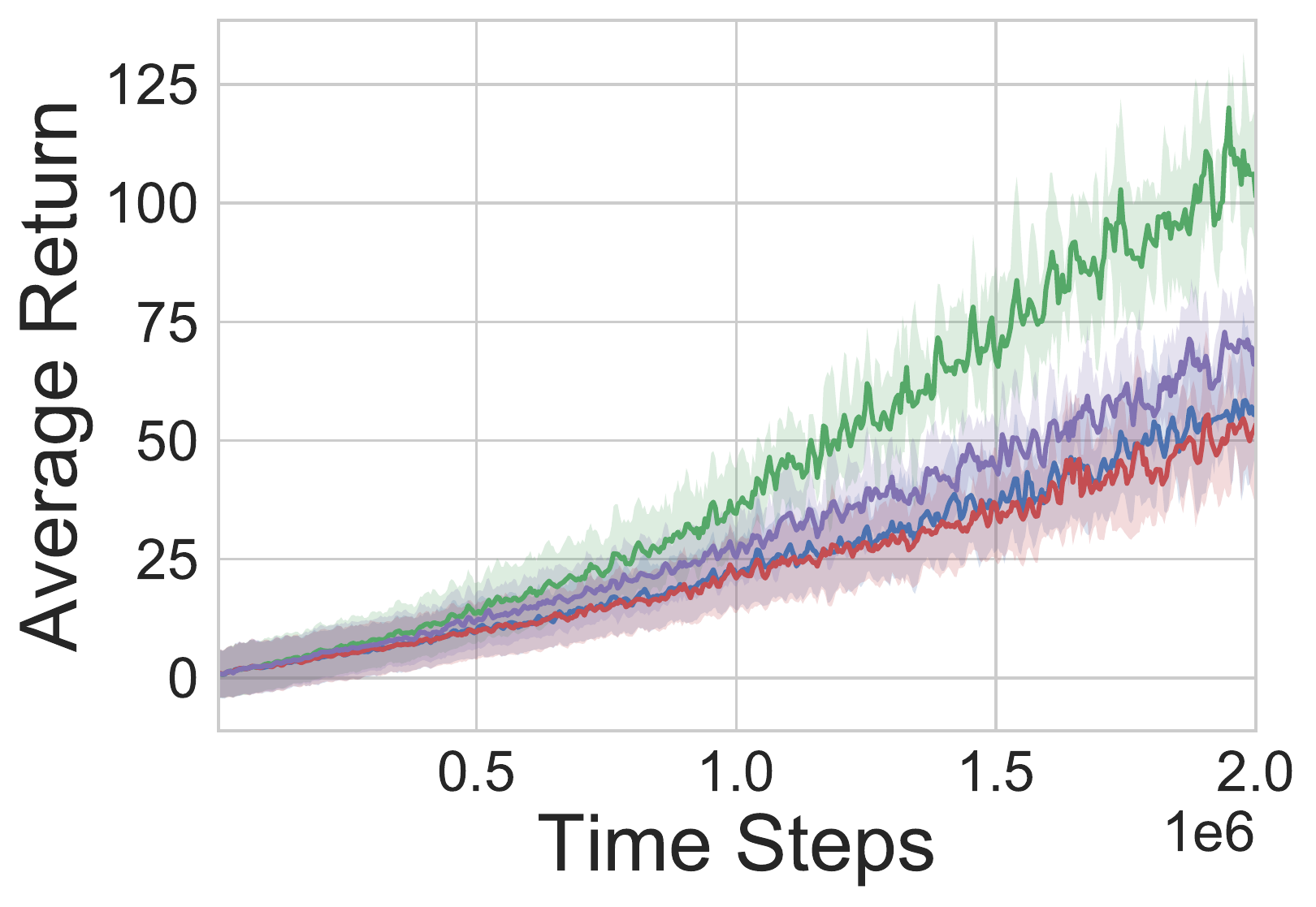}} 
	\vspace{-1.5mm}
	\centering
	\subfigure{\label{fig:legend_ddpg}\includegraphics[width=0.34\linewidth]{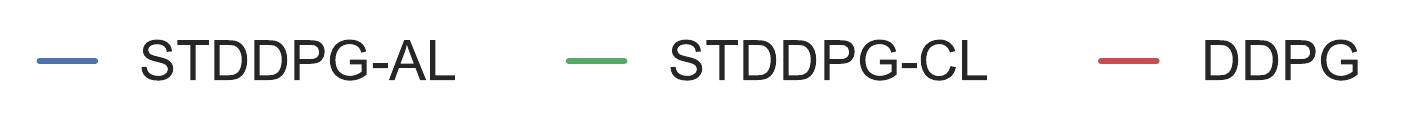}} \\
	\vspace{-4mm}
	\addtocounter{subfigure}{-1}
	\subfigure[${\tt Half Cheetah}$]{\label{fig:ddpg_halfcheetah}\includegraphics[width=0.22\linewidth]{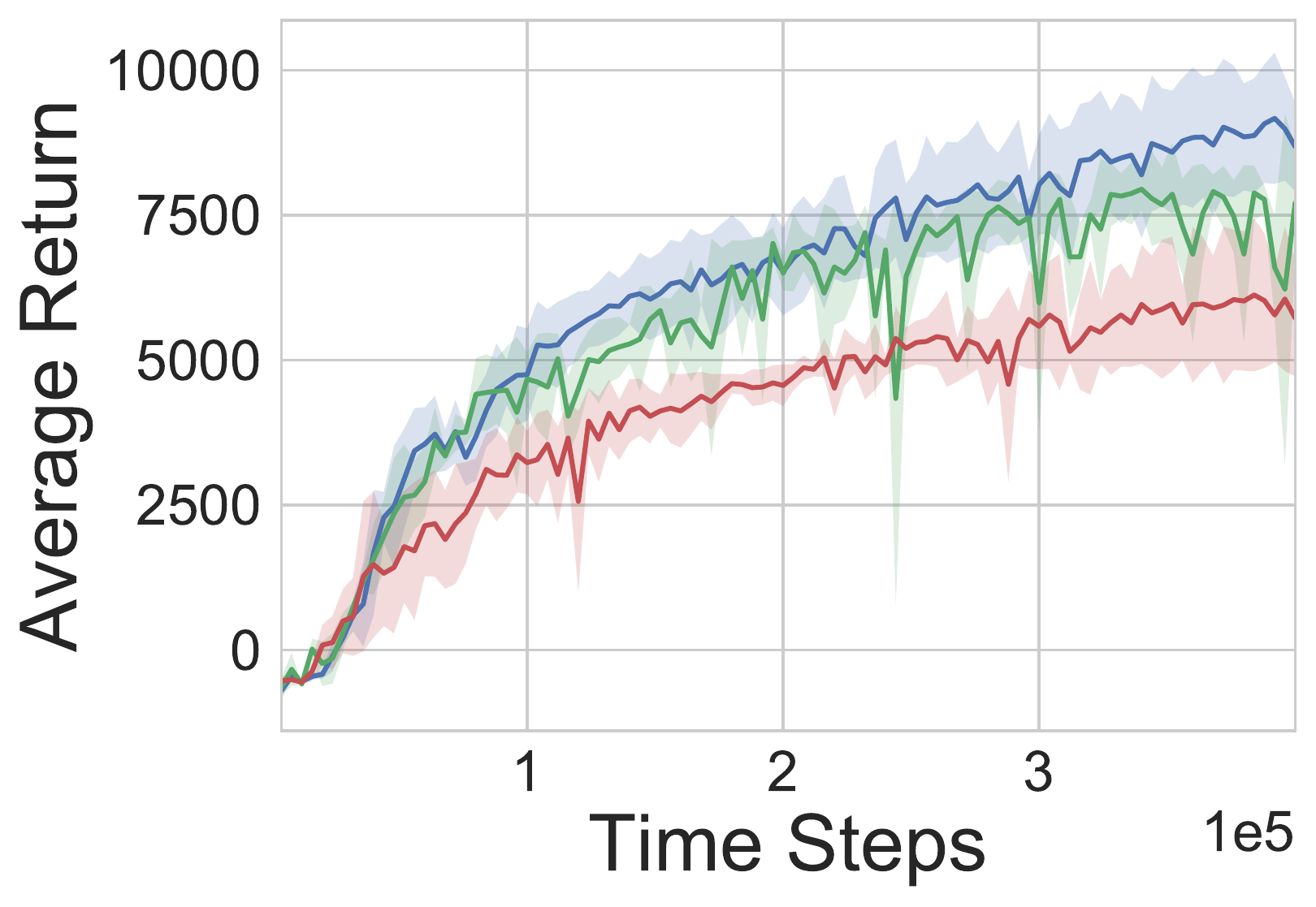}} \hfill
	\subfigure[${\tt Swimmer}$]{\label{fig:ddpg_swimmer}\includegraphics[width=0.22\linewidth]{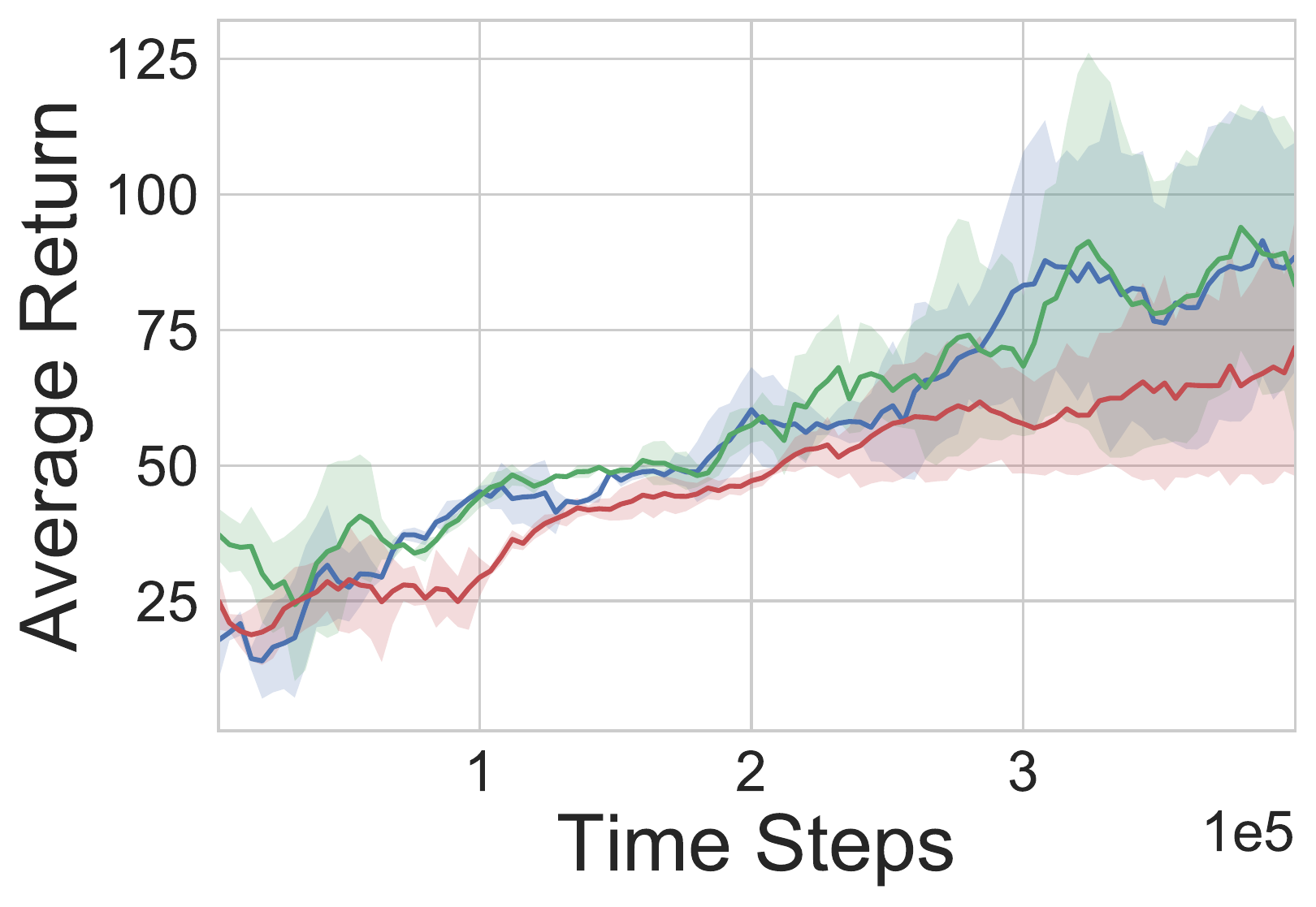}} \hfill 
	\subfigure[${\tt Hopper}$]{\label{fig:ddpg_hopper}\includegraphics[width=0.22\linewidth]{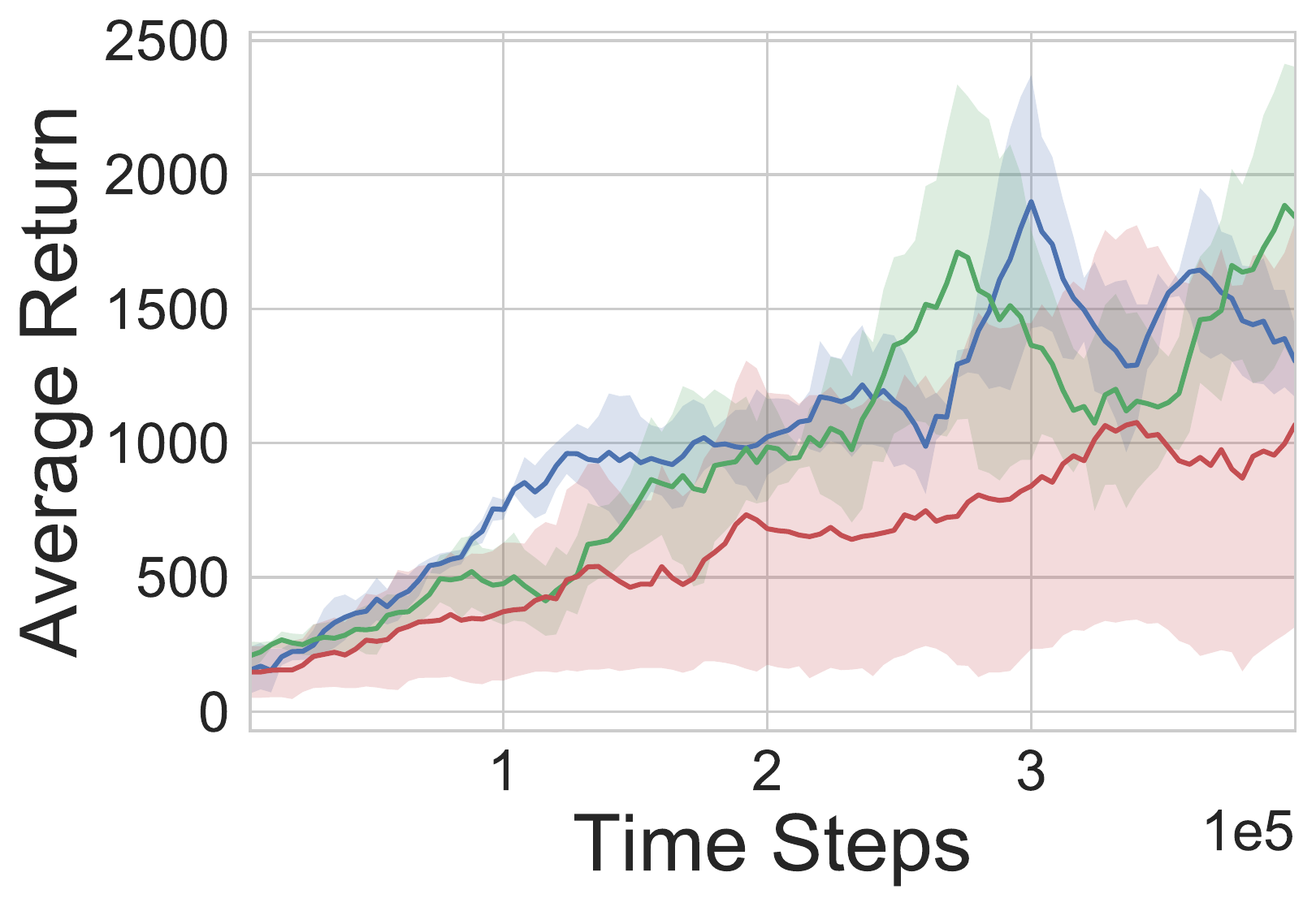}} \hfill
	\subfigure[${\tt Walker2d}$]{\label{fig:ddpg_walker2d}\includegraphics[width=0.22\linewidth]{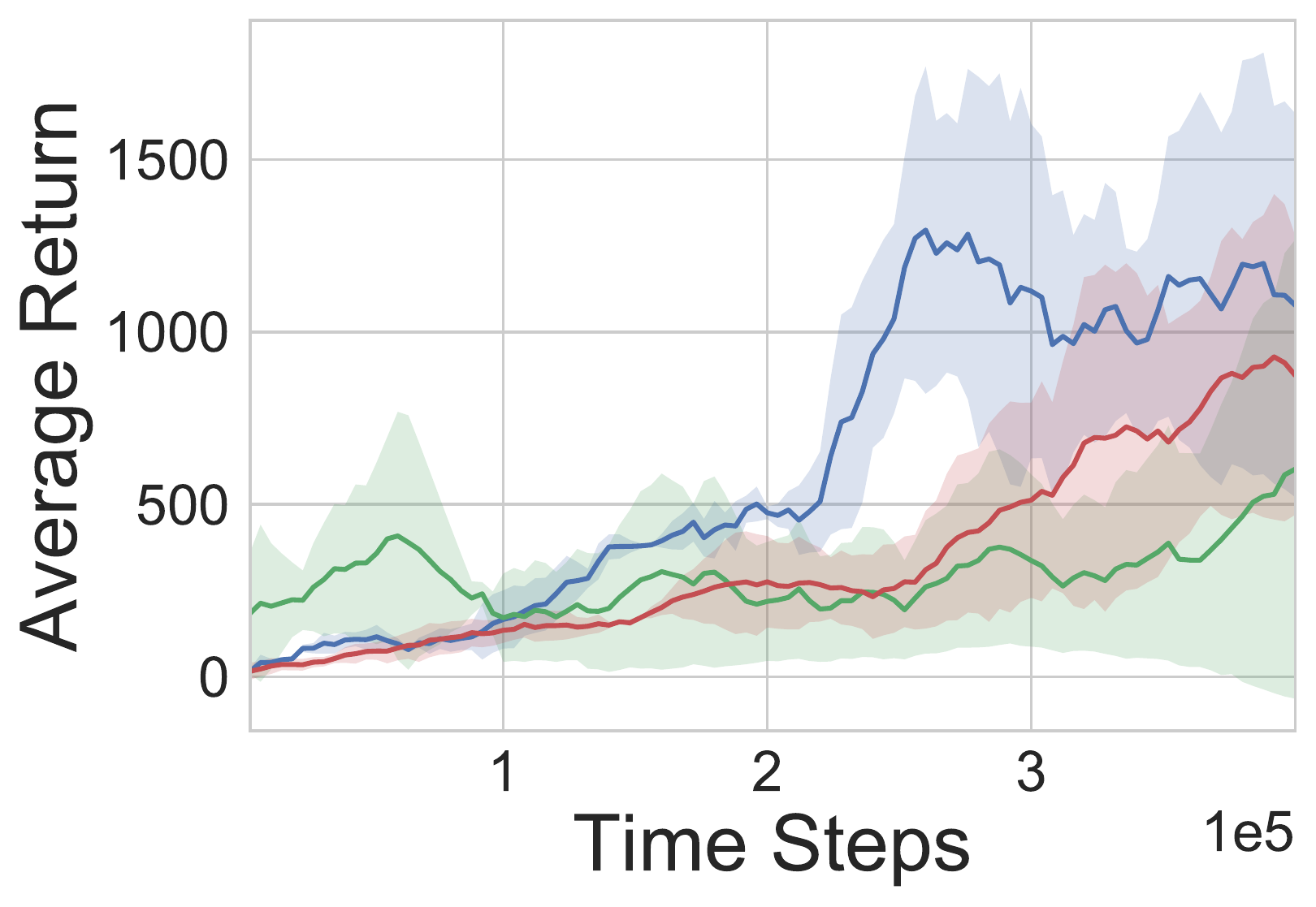}}
	\vspace{-1.5mm}
	\centering
	\subfigure{\label{fig:legend_sac}\includegraphics[width=0.32\linewidth]{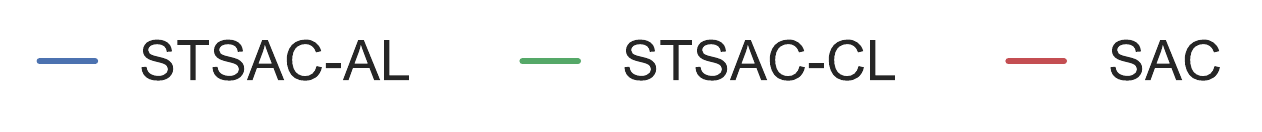}} \\
	\vspace{-4mm}
	\addtocounter{subfigure}{-1}
	\subfigure[${\tt HalfCheetah}$]{\label{fig:sac_halfcheetah}\includegraphics[width=0.22\linewidth]{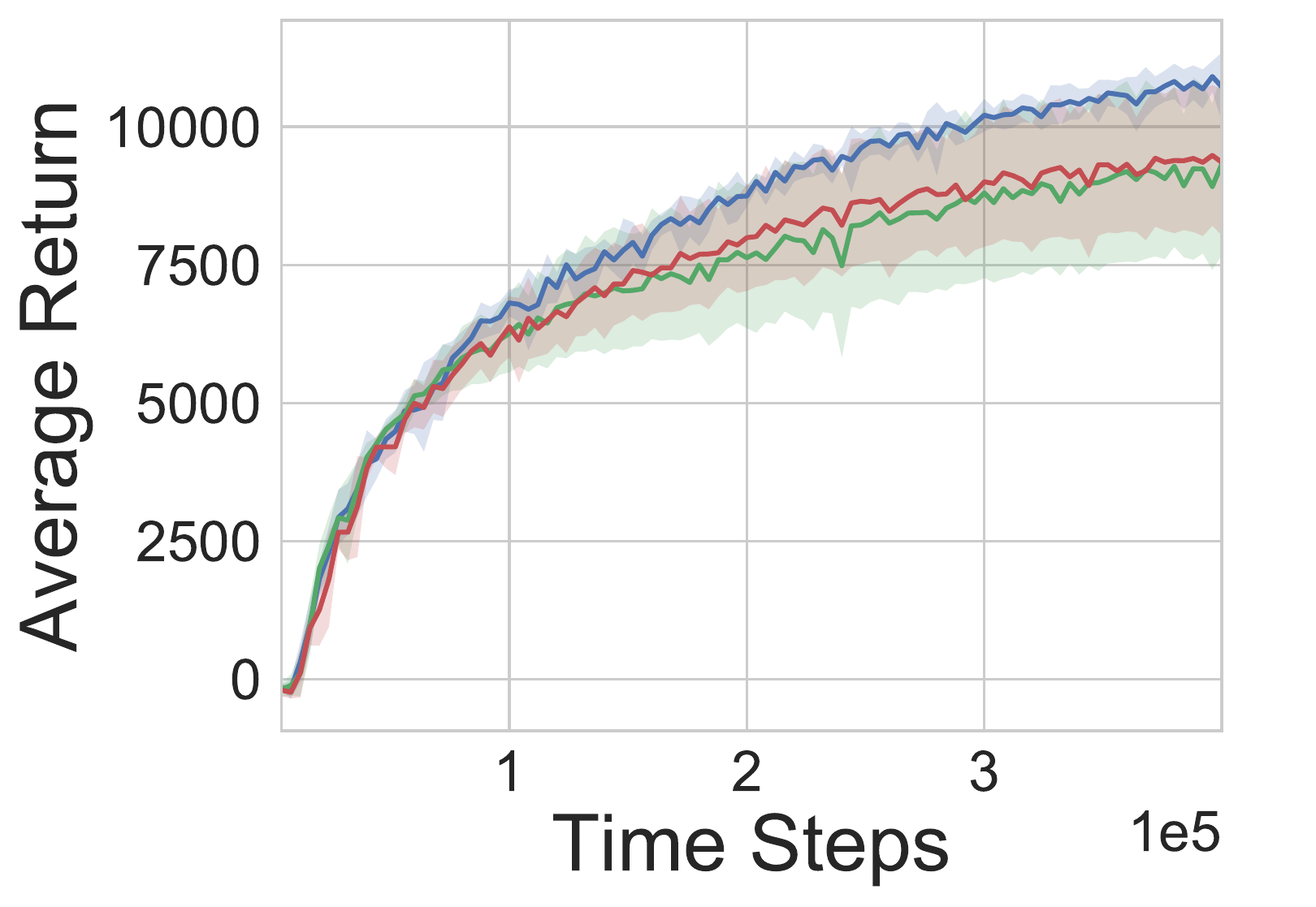}} \hfill
	\subfigure[${\tt Swimmer}$]{\label{fig:sac_swimmer}\includegraphics[width=0.22\linewidth]{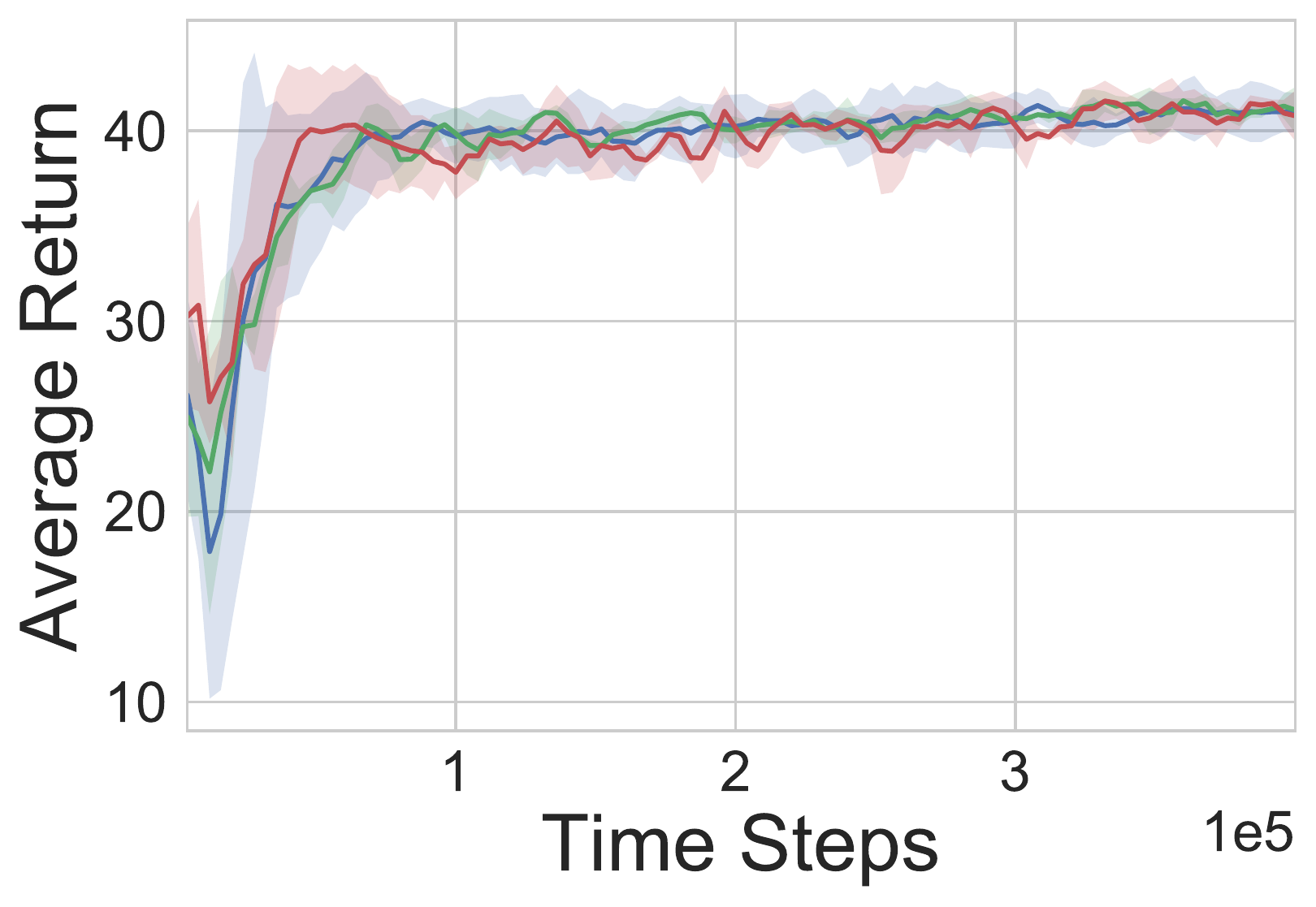}} \hfill 
	\subfigure[${\tt Hopper}$]{\label{fig:sac_hopper}\includegraphics[width=0.22\linewidth]{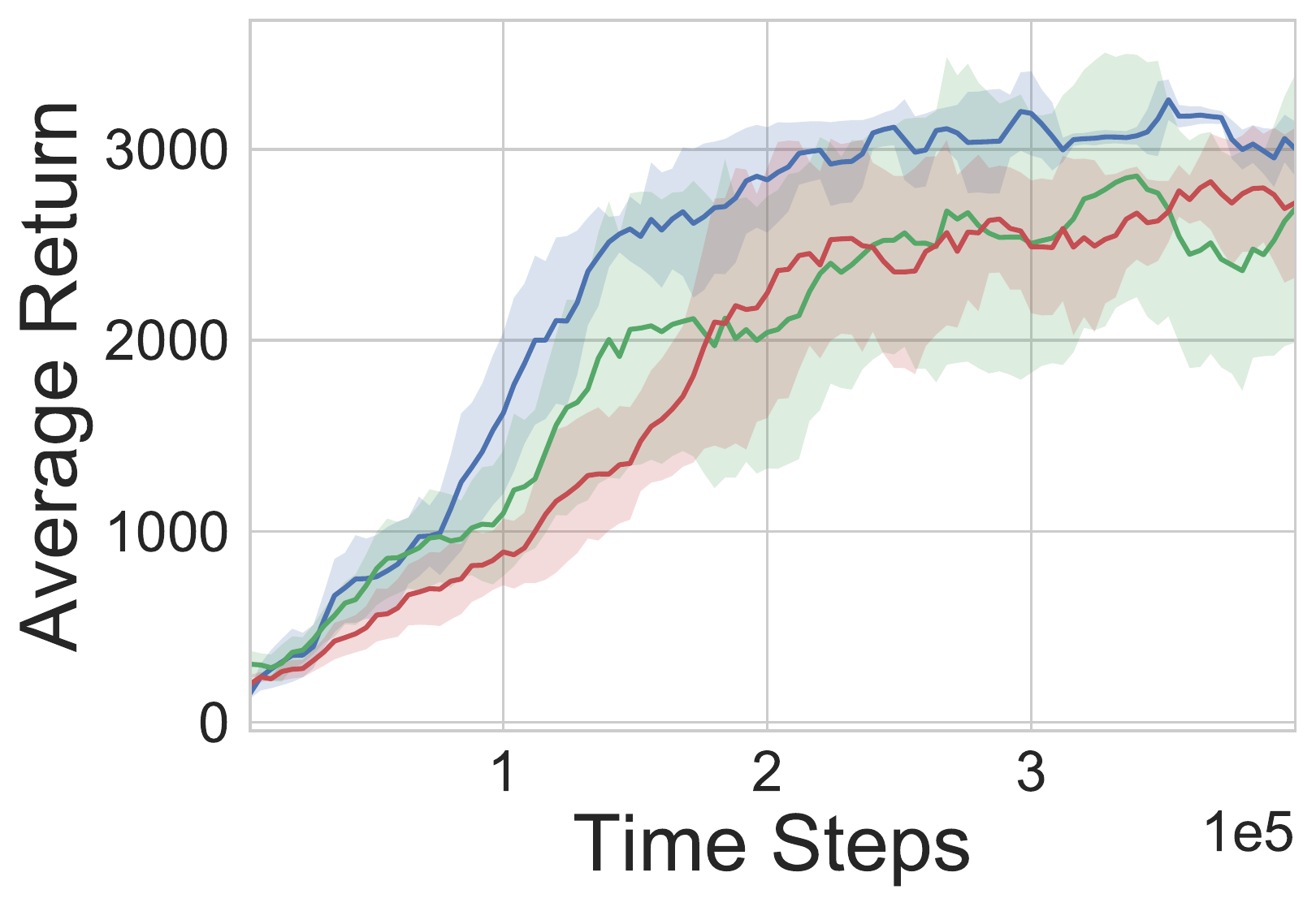}} \hfill
	\subfigure[${\tt Walker2d}$]{\label{fig:sac_walker2d}\includegraphics[width=0.22\linewidth]{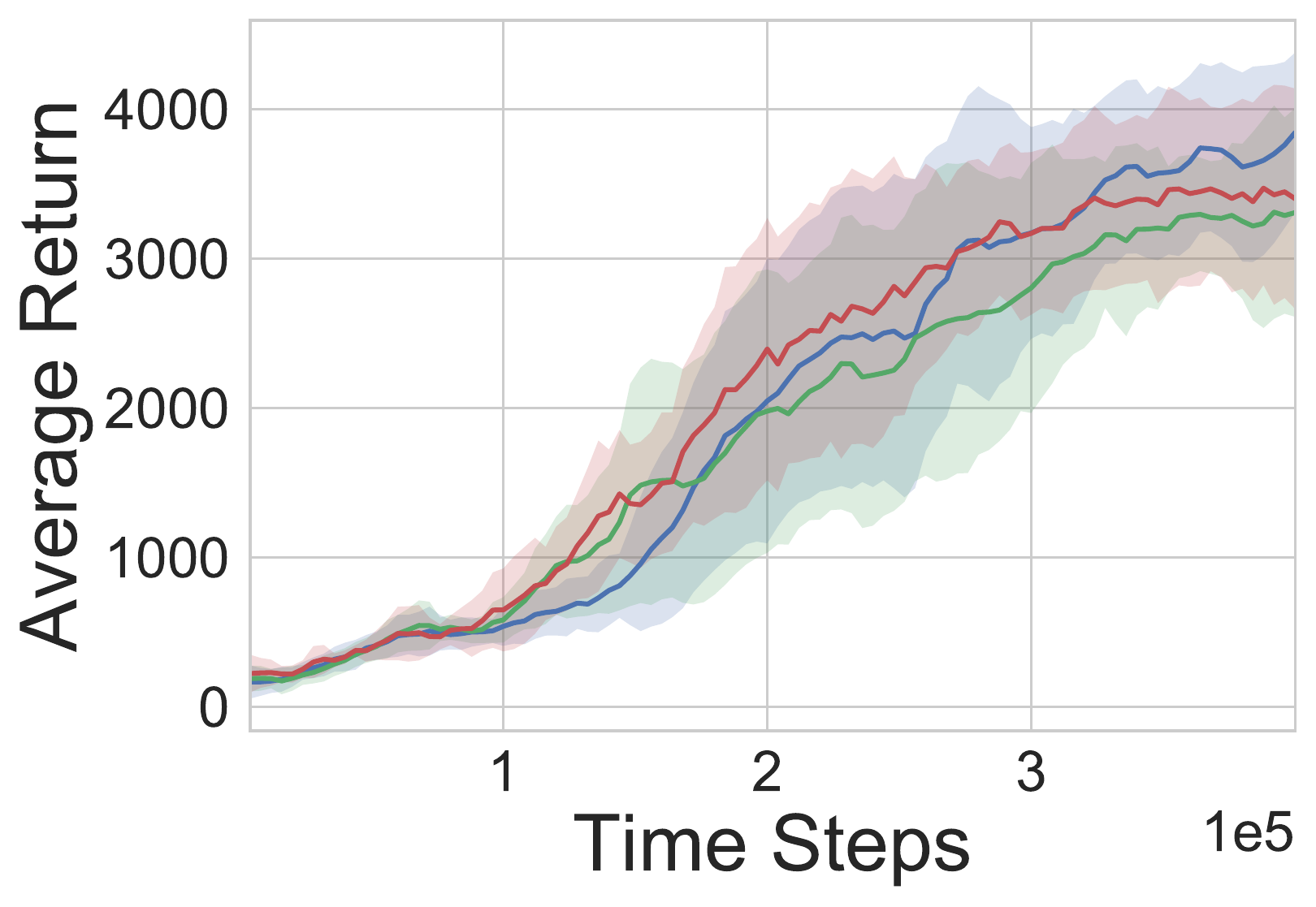}}
	\vspace{-3mm}
	\caption{Comparison of {\ac}, {\ddpg}, {\sac} with their Stackelberg versions on OpenAI gym environments. 
	}
	\label{fig:sac}
\end{figure*}

\section{Experiments}
\label{sec:exp}
We now show the results of extensive experiments comparing the Stackelberg {\act} algorithms with the comparable {\act} algorithms. We find that the {\act} algorithms with the Stackelberg gradient dynamics always perform at least as well and often significantly outperform the standard gradient dynamics. Moreover, we provide game-theoretic interpretations of the results.

We run experiments on the OpenAI gym platform \citep{brockman2016openai} with the Mujoco Physics simulator \citep{todorov2012mujoco}. The performance of each algorithm is evaluated by the average episode return versus the number of time steps (state transitions after taking an action according to the policy). 
For a fair comparison, the hyper-parameters for the actor and critic including the neural network architectures are set equal when comparing the Stackelberg actor-critic algorithms with the stand normal actor-critic algorithms. The implementation details are in Appendix~\ref{exp_detail}, and importantly, the Stackelberg actor-critic algorithms are not significantly more computationally expensive than the normal algorithms.

\paragraph{Performance.}
Figures~\ref{fig:cartpole1}--\ref{fig:walker} show the performance of {\STAC} and {\ac} on several tasks. We also experiment with the common heuristic of ``unrolling'' the critic $m$ steps between actor steps. For each task, {\STAC} with multiple critic unrolling steps performs the best. This is due to the fact when the critic is closer to the best response, then the real response of the critic is closer to what is anticipated by the Stackelberg gradient for the actor. Interestingly, in CartPole, {\STAC} with $m=1$ performs even better than {\ac} with $m=80$.

Figures~\ref{fig:ddpg_halfcheetah}--\ref{fig:ddpg_walker2d} show the performance of {\stddpg}-{\tt \AL} and {\stddpg}-{\tt \CL} in comparison to {\ddpg}. We  observe that on each task, {\stddpg}-{\tt \AL} outperforms {\ddpg} by a clear margin, whereas {\stddpg}-{\tt \CL} has overall better performance than {\ddpg} except on ${\tt Walker2d}$.  Figures~\ref{fig:sac_halfcheetah}--\ref{fig:sac_walker2d} show the performance of {\stsac}-{\tt \AL} and {\stsac}-{\tt \CL} in comparison to {\sac}. 

In all experiments, when the actor is the leader, the Stackelberg versions either outperform or are comparable to the existing {\act} algorithms, offering compelling evidence that the Stackelberg framework has an empirical advantage in many tasks and settings. We now provide game-theoretic interpretations of the experimental results and connect back to the examples and observations from Section~\ref{sec:example}.

\paragraph{Game-Theoretic Interpretations.}
{\sac} is considered the state-of-the-art model-free {\rl} algorithm and we observe it significantly outperforms {\ddpg} (e.g., on ${\tt Hopper}$ and ${\tt Walker2d}$). The common interpretation of its advantage is that {\sac} encourages exploration by penalizing low entropy policies. Here we provide another viewpoint. 

From a game-theoretic perspective, the objective functions of {\ac}  and {\ddpg} take on hidden linear and hidden quadratic structures for the actor and critic. This structure can result in cyclic behavior for individual gradient dynamics as shown in Section~\ref{sec:example}. {\sac} constructs a more well-conditioned game structure by regularizing the actor objective, which leads to the learning dynamics converging more directly to the equilibrium as seen in Section~\ref{sec:example}.
This also explains why we observe improved performance with {\STAC} and {\stddpg}-{\tt \AL} compared to {\ac} and {\ddpg}, but the performance gap between {\stsac}-{\tt \AL} and {\sac} is not as significant. 

Comparing {\tt \AL} with {\tt \CL}, the actor as the leader always outperforms the critic as the leader in our experiments. As described in Section~\ref{sec:example}, the critic objective is typically a quadratic mean square error objective, which results in a hidden quadratic structure, whereas the actor's objective typically has a hidden linear structure due to parameterization of the $Q$ network and policy.  Thus, the critic cost structure is more well-suited for computing an approximate local best response since it is more likely to be well-conditioned, and so the critic as the follower is the more natural hierarchical game structure. Unrolling the critic for multiple steps to approximate this structure and has been  shown to perform well empirically~\citep{schulman2015trust}. Algorithm~\ref{alg:framework_2} (Appendix~\ref{exp_detail}) describes this method for the Stackelberg framework.

\section{Conclusion}
We revisit the standard {\act} algorithms from a game-theoretic perspective to capture the hierarchical interaction structure and introduce a Stackelberg framework for {\act} algorithms. In this framework, we introduce novel Stackelberg versions of existing {\act} algorithms. In experiments on a number of environments, we show that the Stackelberg {\act} algorithms always outperform the existing counterparts when the actor plays the leader. 

\clearpage
\bibliography{reference}

\appendix
\onecolumn

\section{Motivation Example Details} \label{appendix:example}
In this appendix section, we provide more detail for the example in Section~\ref{sec:motivation}. 
\begin{figure*}[!ht]
	\centering
	\subfigure[Individual gradient]{\label{fig:traj_gd_2}\includegraphics[width=0.3\linewidth]{fig/traj_gd_2.pdf}} 
	\subfigure[Stackelberg gradient]{\label{fig:traj_stgd_2}\includegraphics[width=0.3\linewidth]{fig/traj_stgd_2.pdf}}
	\subfigure[Regularized Stackelberg gradient]{\includegraphics[width=0.3\linewidth]{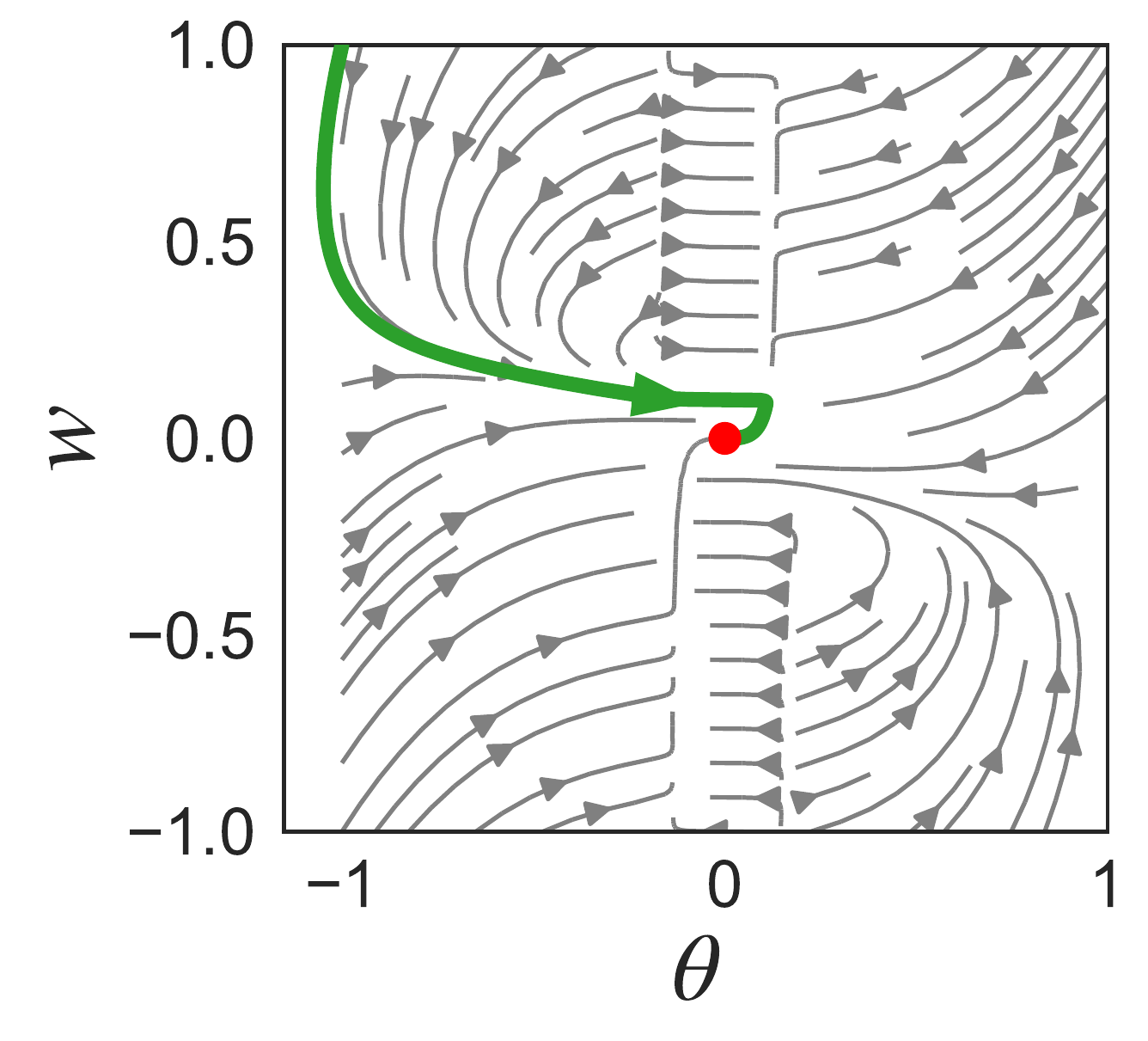}\label{fig:traj_stgd_reg_2}}	\hfill
	\caption{Vector fields and trajectories of the individual gradient, Stackelberg gradient and regularized Stackelberg gradient updates. The Stackelberg updates eliminate cycling by changing the shape of the vector field.}
	\label{fig:examples-a}
\end{figure*}

\begin{figure*}[h!]
	\centering
	\subfigure[Error]{\label{fig:error_2}\includegraphics[width=0.35\linewidth]{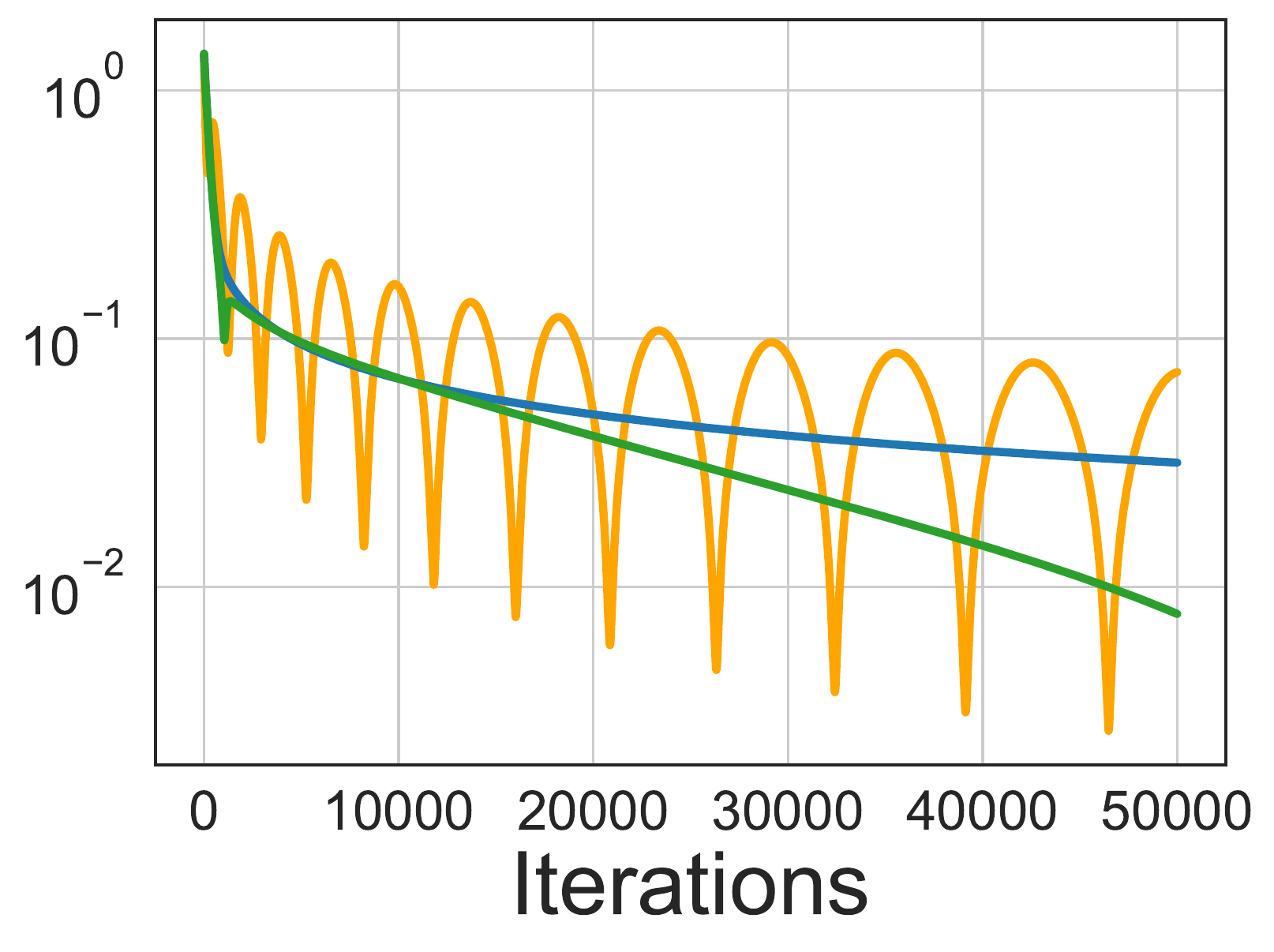}} 
	\subfigure[Return]{\label{fig:return_2}\includegraphics[width=0.35\linewidth]{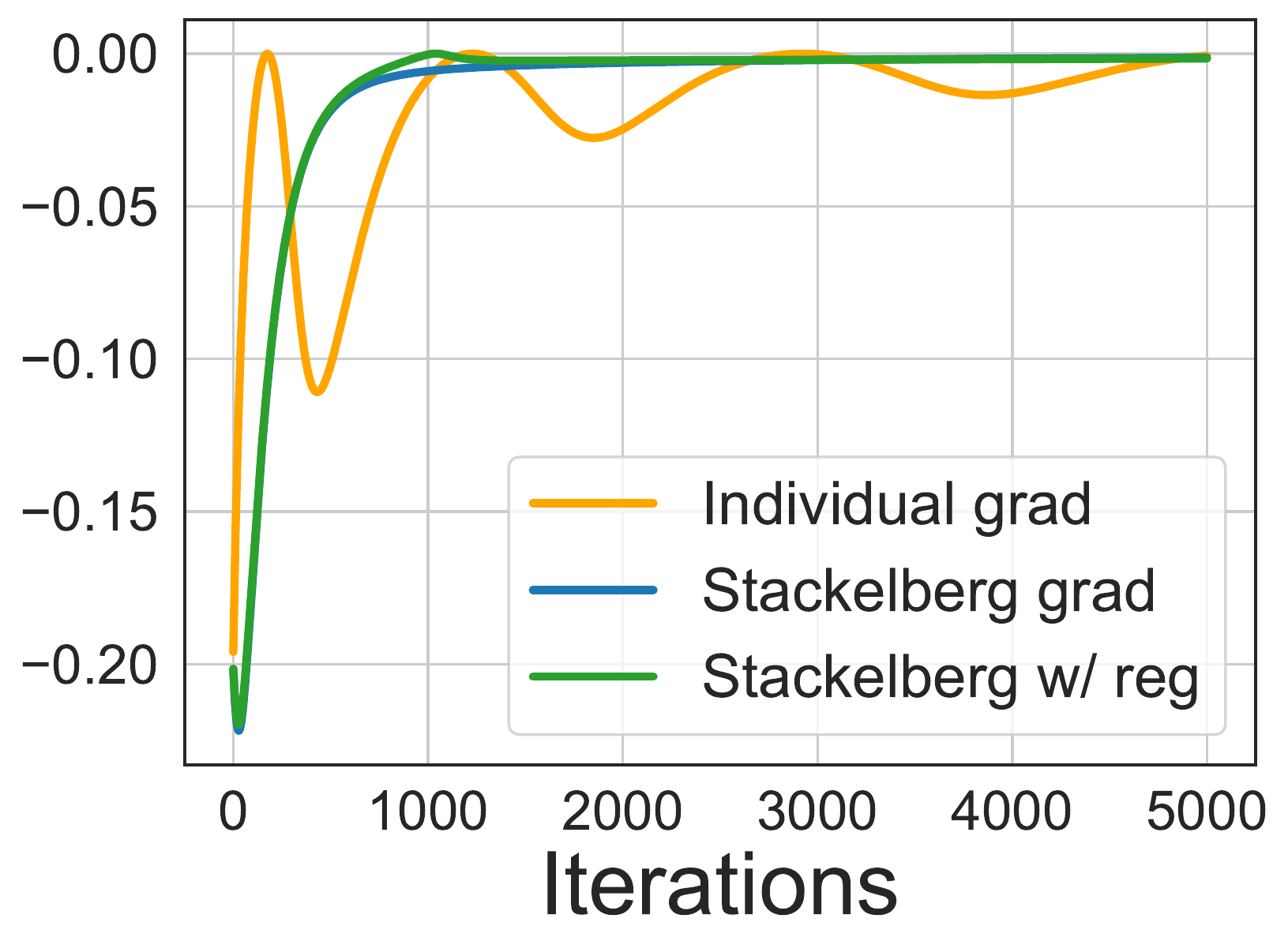}} \hfill 
	\caption{(a) Convergence error $\|w-w^*\|^2 + \|\theta - \theta^*\|^2$ where $(\theta^\ast,w^\ast)=(0,0)$ is the equilibrium. (b) The return $R(\theta)$ of the actor. The Stackelberg update eliminates cycling and hence, converges more directly to the equilibrium as can be seen in (a), whereas the individual gradient update oscillates significantly. Regularization helps to speed up convergence.}
	\label{fig:examples-a2}
\end{figure*}
Recall the motivating example in which the actor plays the leader with the objective function $J(\theta, w) = w\cdot \theta$, and the critic plays the follower with objective function $L(\theta, w) = (w\cdot \theta + \frac{1}{5}\theta^2)^2$.
Figure~\ref{fig:examples-a} shows the vector fields and trajectories of each of the updates: individual gradient play\footnote{In the learning in games literature, this is also often referred to as simultaneous gradient play or simultaneous gradient descent-ascent.}, Stackelberg gradient play, and regularized Stackelberg gradient play.
In Figure~\ref{fig:traj_gd_2}, we observe clear cycling behavior. Such cycling behavior may be an indication of reduced reliability along the learning path and is often exacerbated by noise. Generally speaking, it is more desirable to observe   smooth,
monotonic changes in performance as compared to cycling behavior or noisy fluctuations around a observable trend.  The reason for this is that when we go to deploy such algorithms in the real world, it can be extremely costly to have the algorithm perform in oscillatory or even unpredictable ways. This is in particular true when, as is often the case, there are unmodeled exogenous inputs or environmental factors. 

On the other hand, Stackelberg gradient converges more directly to the equilibrium point $(\theta^*, w^*) = (0, 0)$ and shown in both Figures~\ref{fig:traj_stgd_2} and \ref{fig:traj_stgd_reg_2} where the latter are the trajectories of the regularized Stackelberg gradient introduced in Section~\ref{sec:regularization}. Figure~\ref{fig:error_2}  shows the error $\|w-w^*\|^2 + \|\theta - \theta^*\|^2$ and Figure \ref{fig:return_2} shows the return $R(\theta)$ of each of the updates. We can observe that the cycling is mitigated and convergence accelerated by optimizing using the Stackelberg gradient, which leads to more stable returns along the learning.

\begin{figure*}[h!]
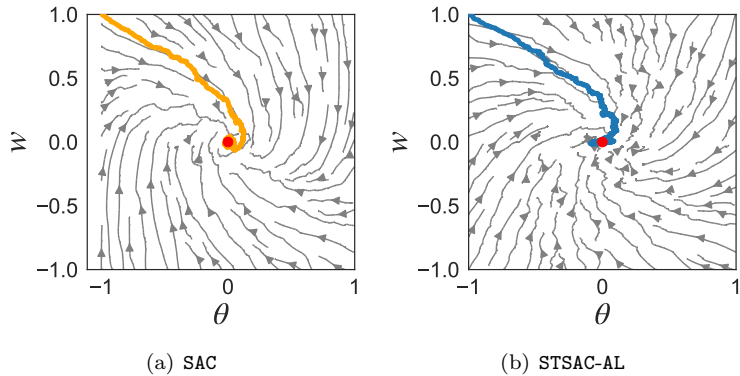

	\centering
	\subfigure[{\sac}]{\label{fig:traj_gd_3}\includegraphics[width=0.3\linewidth]{fig/traj_gd_3.pdf}} \subfigure[{\stsac}-{\tt \AL}]{\label{fig:traj_stgd_3}\includegraphics[width=0.3\linewidth]{fig/traj_stgd_3.pdf}}\hfill
	\caption{Vector fields and trajectories of the {\sac} and {\stsac}-{\tt \AL} updates. }
	\label{fig:examples-b1}
\end{figure*}
\begin{figure*}[h!]
	\centering
	
	\subfigure[Error of {\sac} and {\stsac}-{\tt AL}]{\label{fig:error_3}\includegraphics[width=0.35\linewidth]{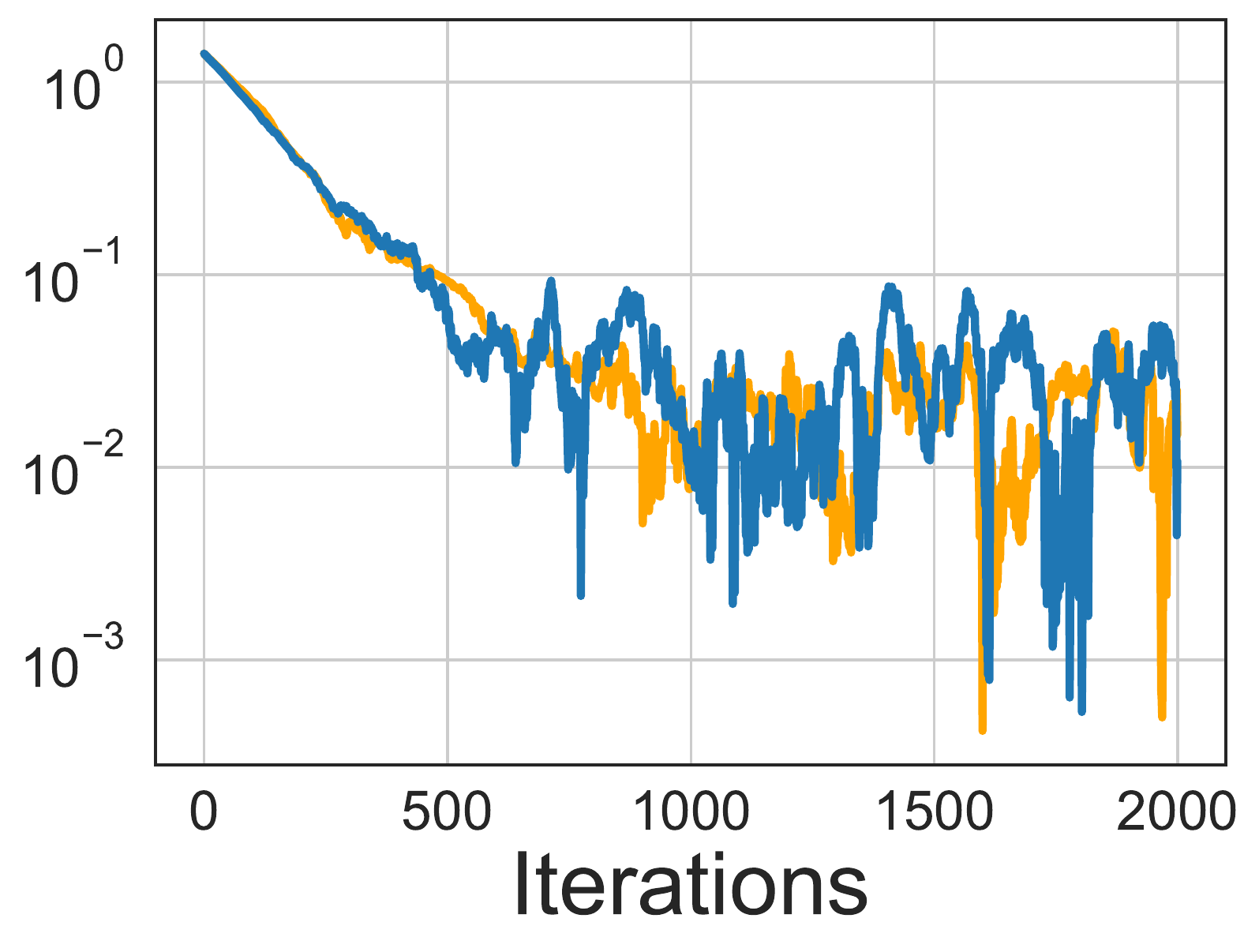}}
	\subfigure[Return of {\sac} and {\stsac}-{\tt \AL}]{\label{fig:return_3}\includegraphics[width=0.35\linewidth]{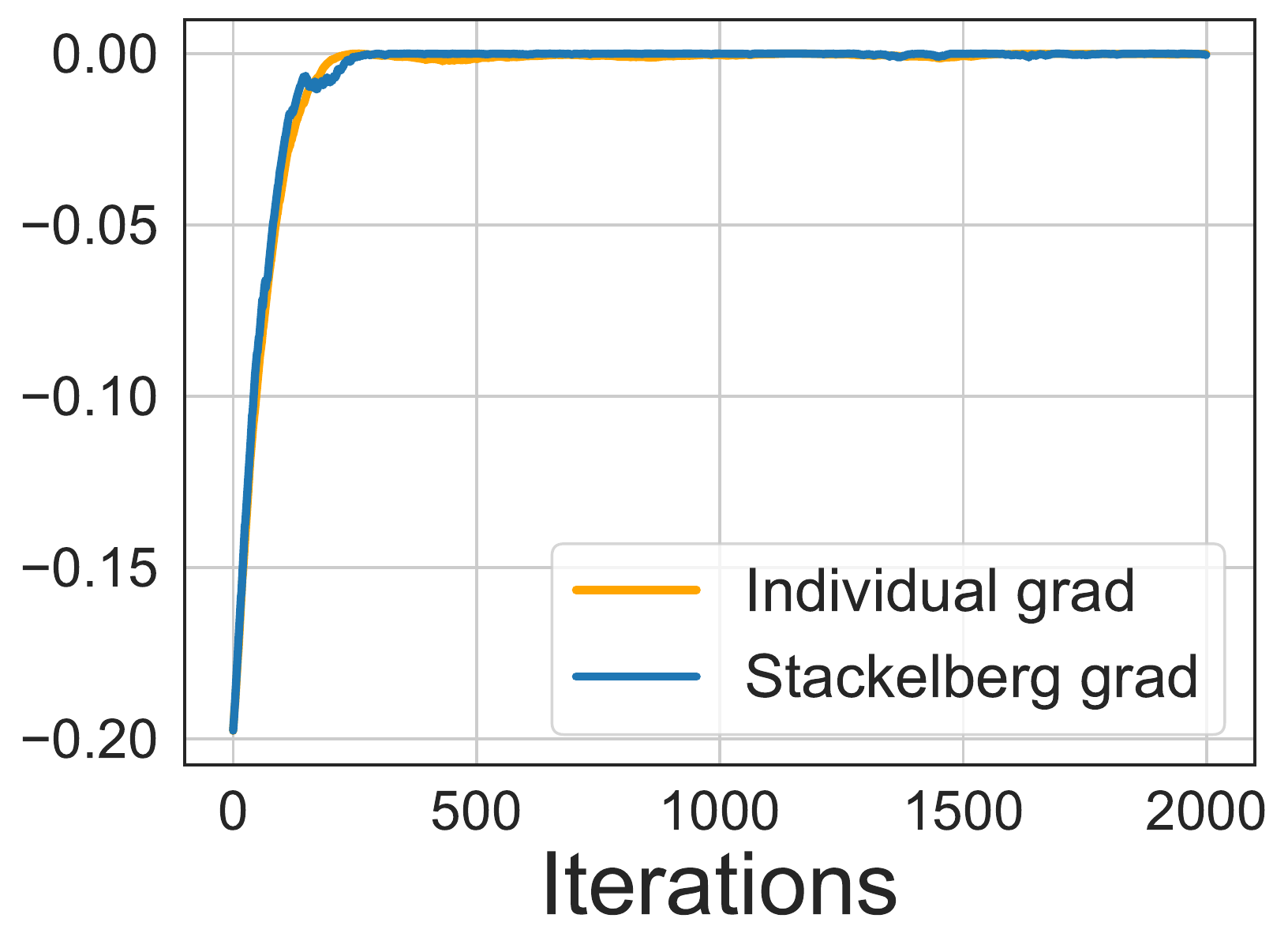}}\hfill
	\caption{ (a) Error for each algorithm, {\sac} and {\stsac}-{\AL}, $\|w-w^*\|^2 + \|\theta - \theta^*\|^2$ where $(\theta^\ast,w^\ast)=(0,0)$ is the equilibrium. (b) Return of the actor $R(\theta)$. }
	\label{fig:examples-b2}
\end{figure*}

In Figures~\ref{fig:examples-b1} and \ref{fig:examples-b2}, we show the result of 
adding entropy regularization to the actor's objective using the {\sac} algorithm. Since {\sac} involves sampling from an stochastic policy, we plot the empirical mean gradient vector fields in Figure~\ref{fig:traj_gd_3} and Figure~\ref{fig:traj_stgd_3}, where the gradients for update are estimated by samples. With the entropy regularization, both gradient updates converge much faster and the gap between them are less significant (Figure~\ref{fig:error_3} and \ref{fig:return_3}).

\section{Proof of Theorem~\ref{THM:STACKPOLICYGRAD}} \label{th1_proof}

Recall that  the critic's objective is given by
$L(\theta,w) = \textstyle \mathbb{E}_{s \sim \rho, a \sim \pi_\theta (\cdot | s)} [ (Q_w(s, a) - Q^\pi(s, a) )^2 ]$.
The derivative is computed as follows:
\begin{align*}
	\nabla_\theta L(\theta, w) & = \nabla_\theta \int_{s_0} \rho(s_0) \int_{a_0} \pi_\theta (a_0|s_0) \left(Q_w(s_0, a_0) - Q^\pi(s_0, a_0) \right)^2 \mathrm{d} a_0 \mathrm{d} s_0 \notag \\
	& =  \int_{s_0} \rho(s_0) \int_{a_0} \nabla_\theta \pi_\theta(a_0|s_0) \left(Q_w(s_0, a_0) - Q^\pi(s_0, a_0) \right)^2 \mathrm{d} a_0 \mathrm{d} s_0 \nonumber \\
	& \quad +  \int_{s_0} \rho(s_0) \int_{a_0} \pi_\theta (a_0|s_0) \nabla_\theta \left(Q_w(s_0, a_0) - Q^\pi(s_0, a_0) \right)^2 \mathrm{d} a_0 \mathrm{d} s_0 \notag \\
	& =  \int_{s_0} \rho(s_0) \int_{a_0} \pi_\theta (a_0|s_0) \nabla_\theta \log \pi_\theta(a_0|s_0) \left(Q_w(s_0, a_0) - Q^\pi(s_0, a_0) \right)^2 \mathrm{d} a_0 \mathrm{d} s_0 \nonumber \\
	& \quad + 2 \int_{s_0} \rho(s_0) \int_{a_0} \pi_\theta (a_0|s_0) \left(Q^\pi(s_0, a_0) - Q_w(s_0, a_0) \right) \nabla_\theta Q^\pi(s_0, a_0) \mathrm{d} a_0 \mathrm{d} s_0. 
\end{align*}
From here, it remains to compute $\nabla_\theta Q^\pi(s_0, a_0)$. To do so, recall that 
$Q^\pi(s_t, a_t)$ and $V^\pi(s_t)$ are given by
\begin{align*}
	Q^\pi(s_t, a_t) &={\textstyle \mathbb{E}_{\tau \sim \pi} \big[ \sum_{t' = t}^T \gamma^{t'-t} r(s_{t'}, a_{t'}) | s_t, a_t \big]}=r(s_t, a_t) + \gamma \int_{s'} P(s'|s_t, a_t) V^\pi(s') \mathrm{d} s',	\end{align*}
and
\begin{align*}
	V^\pi(s_t) & = {\textstyle \mathbb{E}_{\tau \sim \pi} \big[ \sum_{t' = t}^T \gamma^{t'-t} r(s_{t'}, a_{t'}) | s_t \big]}=\int_{a} \pi_\theta(a|s_t) Q^\pi(s_t, a) \mathrm{d} a.
\end{align*}
Hence, $\nabla_\theta Q^\pi(s_0, a_0)$ is computed as follows:
\begin{align}
	& \quad \nabla_\theta Q^\pi(s_0, a_0)  =  \gamma \int_{s_1} P(s_1|s_0, a_0) \nabla_\theta V^\pi(s_1) \mathrm{d} s_1 \notag \\
	& =  \gamma \int_{s_1} P(s_1|s_0, a_0) \int_{a_1} \left( \nabla_\theta \pi_\theta(a_1|s_1) Q^\pi(s_1,a_1) +\pi_\theta(a_1|s_1) \nabla_\theta Q^\pi(s_1,a_1) \right) \mathrm{d}a_1 \mathrm{d}s_1 \notag \\
	& = \gamma \int_{s_1} P(s_1|s_0, a_0) \int_{a_1} \pi_\theta(a_1|s_1) \nabla_\theta \log \pi_\theta(a_1|s_1) Q^\pi(s_1,a_1) \mathrm{d}a_1 \mathrm{d}s_1 \nonumber \\
	& \quad + \gamma^2 \int_{s_1} P(s_1|s_0, a_0) \int_{a_1} \pi_\theta(a_1|s_1) \int_{s_2} P(s_2|s_1, a_1) \nabla_\theta V^\pi(s_2) \mathrm{d}s_2 \mathrm{d}a_1 \mathrm{d}s_1  \notag \\
	& = \gamma \int_{s_1} P(s_1|s_0, a_0) \int_{a_1} \pi_\theta(a_1|s_1) \nabla_\theta \log \pi_\theta(a_1|s_1) Q^\pi(s_1,a_1) \mathrm{d}a_1 \mathrm{d}s_1 \nonumber \\
	& \quad + \gamma^2 \int_{s_1} P(s_1|s_0, a_0) \int_{a_1} \pi_\theta(a_1|s_1) \int_{s_2} P(s_2|s_1, a_1) \int_{a_2} \pi_\theta(a_2|s_2) \nabla_\theta \log \pi_\theta(a_2|s_2) Q^\pi(s_2, a_2)  \mathrm{d}a_2 \mathrm{d}s_2 \mathrm{d}a_1 \mathrm{d}s_1 \nonumber \\
	& \quad + \gamma^3 \int_{s_1} P(s_1|s_0, a_0) \int_{a_1} \pi_\theta(a_1|s_1) \int_{s_2} P(s_2|s_1, a_1) \int_{a_2} \pi_\theta(a_2|s_2) \int_{s_3} P(s_3|s_2, a_2) \nabla_\theta V^\pi(s_3) \mathrm{d}s_3  \mathrm{d}a_2 \mathrm{d}s_2 \mathrm{d}a_1 \mathrm{d}s_1 \notag \\
	& = \gamma \int_{\tau} p(\tau_{1:1}|\theta) \nabla_\theta \log \pi_\theta(a_1|s_1) Q^\pi(s_1,a_1) \mathrm{d} \tau_{1:1} \nonumber \\
	& \quad + \gamma^2 \int_{\tau} p(\tau_{1:2}|\theta) \nabla_\theta \log \pi_\theta(a_2|s_2) Q^\pi(s_2,a_2) \mathrm{d} \tau_{1:2} \nonumber \\
	& \quad + \dots \notag \\
	& = \int_{\tau} \sum_{t=1}^T \gamma^t p(\tau_{1:t}|\theta)\nabla_\theta \log \pi_\theta(a_t|s_t) Q^\pi(s_t,a_t) \mathrm{d} \tau. \label{eq:q_grad}
\end{align}
where the last equality is obtained by unrolling and marginalization for the entire length of the trajectory. 

Thus, coming back to the computation of $\nabla_\theta L(\theta,w)$, we have that
\begin{align}
	\nabla_\theta L(\theta, w)
	& =  \int_{s_0} \rho(s_0) \int_{a_0} \pi_\theta (a_0|s_0) \nabla_\theta \log \pi_\theta(a_0|s_0) \left(Q_w(s_0, a_0) - Q^\pi(s_0, a_0) \right)^2 \mathrm{d} a_0 \mathrm{d} s_0 \nonumber \\
	& \quad + 2 \int_{s_0} \rho(s_0) \int_{a_0} \pi_\theta (a_0|s_0) \left(Q^\pi(s_0, a_0) - Q_w(s_0, a_0) \right) \nabla_\theta Q^\pi(s_0, a_0) \mathrm{d} a_0 \mathrm{d} s_0 \notag \\
	& =  \int_{\tau} p(\tau_{0}|\theta) \nabla_\theta \log \pi_\theta(a_0|s_0) \left(Q_w(s_0, a_0) - Q^\pi(s_0, a_0) \right)^2 \nonumber \\
	& \quad + 2
	\sum_{t=1}^T \gamma^t p(\tau_{0:t}|\theta)\nabla_\theta \log \pi_\theta(a_t|s_t) \left(Q^\pi(s_0, a_0) - Q_w(s_0, a_0) \right) Q^\pi(s_t,a_t) \mathrm{d} \tau \notag \\
	& = \mathbb{E}_{\tau \sim \pi_\theta} \left[ \vphantom{\sum_{t=1}^T} \nabla_\theta \log \pi_\theta(a_0|s_0) \left(Q_w(s_0, a_0) - Q^\pi(s_0, a_0) \right)^2 \right. \nonumber \\
	& \qquad\qquad\quad \left. + \sum_{t=1}^T \gamma^t \nabla_\theta \log \pi_\theta(a_t|s_t) \left(Q^\pi(s_0, a_0) - Q_w(s_0, a_0) \right) Q^\pi(s_t,a_t) \right]\notag
\end{align}
which completes the proof.

\section{Proof of Proposition~\ref{PROP:STACKVALUEGRAD}} \label{prop1_proof}
The critic's objective is given by $L(\theta, w) =  \mathbb{E}_{s \sim \rho} \left[ \left(V_w(s) - V^\pi(s) \right)^2 \right]$. Hence, taking the derivative with respect to $\theta$, we have that
\begin{align}
	\nabla_\theta L(\theta, w)
	& = \int_{s_0} \rho(s_0) \nabla_\theta (V_w(s_0) - V^\pi(s_0))^2 \mathrm{d} s_0 \notag \\
	& = 2  \int_{s_0} \rho(s_0) (V^\pi(s_0) - V_w(s_0)) \nabla_\theta V^\pi(s_0) \mathrm{d} s_0. \label{eq:proof2}
\end{align}
Now we compute $\nabla_\theta V^\pi(s_0)$ in~\eqref{eq:proof2}. Use the result of~\eqref{eq:q_grad}, we have
\begin{align}
	\nabla_\theta V^\pi(s_0) & = \int_{a_0}  \nabla_\theta \pi_\theta(a_0|s_0) Q^\pi(s_0,a_0) +\pi_\theta(a_0|s_0) \nabla_\theta Q^\pi(s_0,a_0)  \mathrm{d}a_0 \notag \\
	&=  \int_{\tau} \pi_\theta(a_0|s_0) \left( \nabla_\theta \log \pi_\theta(a_0|s_0) Q^\pi(s_0,a_0) + \sum_{t=1}^T \gamma^t p(\tau_{1:t}|\theta)\nabla_\theta \log \pi_\theta(a_t|s_t) Q^\pi(s_t,a_t) \right) \mathrm{d} \tau. \label{eq:v_grad}
\end{align}
Substituting~\eqref{eq:v_grad} into~\eqref{eq:proof2}, we have that
\begin{align}
	\nabla_\theta L(\theta, w) & = 2 \int_{\tau} \sum_{t=0}^T \gamma^t p(\tau_{0:t}|\theta) \nabla_\theta \log \pi_\theta(a_t|s_t) \left(V^\pi(s_0) - V_w(s_0) \right) Q^\pi(s_t, a_t) \mathrm{d} \tau \notag \\
	& = \mathbb{E}_{\tau \sim \pi_\theta} \left[ 2 \sum_{t=0}^T \gamma^t \nabla_\theta \log \pi_\theta(a_t|s_t) \left(V^\pi(s_0) - V_w(s_0) \right) Q^\pi(s_t, a_t) \right] \notag 
\end{align}
which completes the proof.

\section{Proof of Theorem~\ref{THM:CONVERGENCE}}
\label{app_sec:convergenceproof}

Without loss of generality, the actor plays the role of the leader. Consider a differential Stackelberg equilibrium of the game $(\theta^\ast,w^\ast)$ which is locally asymptotically stable\footnote{That is, the local linearization of the above dynamics around the point $(\theta^\ast,w^\ast)$ are in the open left-half complex plane.} for the continuous time dynamical system
\[\begin{bmatrix}\dot{\theta}\\\dot{w}\end{bmatrix}=\begin{bmatrix}\nabla J(\theta,w)\\-\nabla_w L(\theta,w))\end{bmatrix}\]
where
the total derivative of actor in the Stackelberg gradient is given by
\begin{equation*}
	\nabla J(\theta,w) =\nabla_\theta J(\theta, w) - \nabla_{w\theta}^\top L(\theta,w)(\nabla_w^2L(\theta,w))^{-1}\nabla_w J(\theta,w). 
\end{equation*}
and the individual gradient for the critic is $\nabla_w L(\theta, w)$.  The actor and critic employ the discrete time updates given in Algorithm~\ref{alg:framework} where the actor is the leader. Since the actor and critic have unbiased estimates of their gradients 
and the learning rates are chosen as stated in Section~\ref{sec:converge}, then the result of the theorem follows from Theorem 7 in~\citep{fiez2020implicit}. That is, from an initial point $(\theta_0,w_0)\in U$,  the Stackelberg gradient dynamics converge asymptotically to $(\theta^\ast,w^\ast)\in U$ almost surely. 

Indeed, the result holds by the following reasoning. Under the assumptions on the noise processes and stepsize sequences, we  treat the updates in Algorithm~\ref{alg:framework} as a stochastic approximation process $(\theta_{k},w_k)$. Then, we define asymptotic pseudo-trajectories---i.e., linear interpolations  between iterates $(\theta_{k},w_{k})$ and $(\theta_{k+1},w_{k+1})$.  Since $(\theta^\ast,w^\ast)$ is locally asymptotically stable, there exists a neighborhood of $(\theta^\ast,w^\ast)$ and a local Lyapunov function on that neighborhood. This Lyapunov function can be used to show that the continuous time flow also starting from iterates $(\theta_k,w_k)$ and the asymptotic pseudo-trajectories are contracting onto one another asymptotically, for any sequence of iterates starting at $(\theta_0,w_0)\in U$.
Hence, the iterates $(\theta_{k},w_{k})$, in turn, converge asymptotically to $(\theta^\ast,w^\ast)$ almost surely.

\paragraph{Comments on designing gradient estimators.} Methods such as REINFORCE (or Monte Carlo method) provide an unbiased estimator of the follower's individual gradient. Obtaining an unbiased estimate of the total derivative for the leader, on the other hand, is a bit more nuanced. This is because there are multiple gradients being multiplied by one another in the expectation. However, as a heuristic, one way to  approximate it is using the expected value of each of the terms that shows up in the total derivative. 

Depending on the {\act} algorithm and objective functions, following either Theorem~\ref{THM:STACKPOLICYGRAD} (Proposition~\ref{PROP:STACKVALUEGRAD}) or direct derivatives, each term in the total derivative can be computed as an expectation over a  distribution of state and action (generated by current policy in {\ac} and any arbitrary policy in {\ddpg} and {\sac}). 
Take {\ddpg} as an example where
$J(\theta, w) = \mathbb{E}_{\xi \sim \mathcal{D}} \left[ Q_w(s, \mu_\theta(s)) \right]$,
and $L(\theta, w) =  \mathbb{E}_{\xi \sim \mathcal{D}} \left[ \left(Q_w(s, a) - (r + \gamma \Qtar(s', \mu_\theta(s'))) \right)^2 \right]$. The second term in total derivative appears to be a multiplication of several expectations:
\begin{align*}
	\nabla J(\theta,w) & =\nabla_\theta J(\theta, w) - \nabla_{w\theta}^\top L(\theta,w)(\nabla_w^2L(\theta,w))^{-1}\nabla_w J(\theta,w) \nonumber\\
	&=\mathbb{E}_{\xi \sim \mathcal{D}} \left[ \nabla_\theta Q_w(s, \mu_\theta(s)) \right] - \mathbb{E}_{\xi \sim \mathcal{D}} \left[ \nabla_{w\theta} \left(\left(Q_w(s, a) - (r + \gamma \Qtar(s', \mu_\theta(s'))) \right)^2\right) ^\top \right.\nonumber\\
	& \quad \left.\left(   \nabla_w^2 \left(\left(Q_w(s, a) - (r + \gamma \Qtar(s', \mu_\theta(s'))) \right)^2\right) \right)^{-1}  \nabla_w Q_w(s, \mu_\theta(s)) \right]\\
	& \approx  \mathbb{E}_{\xi \sim \mathcal{D}} \left[ \nabla_\theta Q_w(s, \mu_\theta(s)) \right] - \mathbb{E}_{\xi \sim \mathcal{D}} \left[ \nabla_{w\theta} \left(\left(Q_w(s, a) - (r + \gamma \Qtar(s', \mu_\theta(s'))) \right)^2\right) \right]^\top \nonumber\\
	& \quad \left( \mathbb{E}_{\xi \sim \mathcal{D}} \left[ \nabla_w^2 \left(\left(Q_w(s, a) - (r + \gamma \Qtar(s', \mu_\theta(s'))) \right)^2\right) \right] \right)^{-1} \mathbb{E}_{\xi \sim \mathcal{D}} \left[ \nabla_w Q_w(s, \mu_\theta(s)) \right]. 
\end{align*}
For this approximation, we can obtain an unbiased estimate by resetting the simulator as described in \cite[Chapter 11]{sutton2000policy} to estimate each term in the product of expectations. As a result, this is a reasonable heuristic in practice for an approximation to the total derivative. Our policy gradient theorems also provide us a way to derive the estimates of each of these individual terms.
Obtaining unbiased estimates as an active area of research (see, e.g., \citealt{hong2020two,ramponi2021newton}). Moreover, from both a theoretical and practical perspective, understanding how  the batch size affects the estimate of follower Hessian and the total derivative remains open.

\section{Implementation Details} \label{exp_detail}

This section includes complete details about our experiments. Our implementation is developed based on public resource Spinning Up\footnote{Developed by Josh Achiam in 2018: \url{https://spinningup.openai.com/en/latest/}} and our source code is available at \url{https://github.com/LeoZhengZLY/stackelberg-actor-critic-algos}.

We follow the default neural network architecture used in Spinning Up. Particularly, the {\ac} and {\STAC} use networks of size (64, 32) with ${\tt tanh}$ units for both the policy and the value function. The {\ddpg}, {\stddpg}, {\sac}, and {\stsac} use networks of size (256, 256) with ${\tt relu}$ units. The {\ac} and {\STAC} collected 4000 steps of agent-environment interaction per batch and use vanilla gradient descent optimizer and the {\ddpg}, {\stddpg}, {\sac}, and {\stsac} use Adam optimizer with mini-batches of size 100 at each gradient descent step.

The policy gradient terms for {\ac} and {\STAC} are estimated by generalized average estimator (GAE)~\citep{schulman2015high} and critics are updated by Monte Carlo method~\citep{sutton2018reinforcement}. In discrete control task (${\tt CartPole}$), we set the Hessian regularization hyper-parameter $\lambda=0$, and in continuous control tasks (others), we set the regularization hyper-parameter $\lambda=500$.

The performances for {\ac} and {\STAC} are measured as the average trajectory return across the batch collected at each epoch. Performances for {\ddpg}, {\stddpg}, {\sac}, and {\stsac} are measured once every $10,000$ steps by running the deterministic policy (or, in the case of {\sac}, the mean policy) without action noise for ten trajectories, and reporting the average return over those test trajectories.

In our Stackelberg framework, the learning rule for the leader involves computing an inverse-Hessian-vector product for the $\nabla^2_{2} f_2(x_1, x_2)$ inverse term and Jacobian-vector product for the $\nabla_{12} f_2(x_1, x_2)$ terms. The second term can be computed directly by ${\tt autograd.grad}$ in ${\tt torch}$. For the inverse-Hessian-vector term, we implement the conjugate gradient (CG) method using ${\tt autograd.grad}$ iteratively. This enable us to compute and estimate the total derivative on GPU directly and perform Stackelberg gradient update. 
Each CG iteration requires a Hessian vector product (HVP). HVPs can be computed in $\sim 1.5$ times the cost of a gradient~\citep{pearlmutter1994fast}, so the leader update with $k$ CG iterations only costs $\sim 1.5\cdot k$ times a normal gradient. We run CG with $k=10$ so the leader update costs $\sim 15$ times a normal gradient. CG has been applied widely in machine learning~\citep{martens2010deep} and recently at scale for meta-learning~\citep{rajeswaran2019meta} and GANs~\citep{fiez2020implicit}. As observed in~\citep{rajeswaran2019meta, fiez2020implicit}, often $k=5$ in CG is sufficient to get within numerical precision, so we could have had the leader update cost $\sim 7.5$ times a normal gradient. 
In all our experiments, the Stackelberg versions of {\act} algorithms roughly take twice the time to train. This is because the bottleneck in {\rl} is sampling trajectories from the environment rather than gradient computing. This additional time of Stackelberg algorithms would go down if we used $k=5$ in CG. Hence, Stackelberg versions of {\act} algorithms training is not significantly slower normal {\act} algorithms. 

In Algorithm~\ref{alg:framework_2}, we provide a more detailed version of our Stackelberg actor-critic algorithm framework when multiple follower unrolling steps and implicit map regularization are involved.

\begin{algorithm}[t]
	\SetAlgoLined
	\KwIn{{\act} algorithm ${\tt ALG}$, player designations, follower unrolling steps $m$, regularization hyperparameter $\lambda$, and learning rate sequences $\alpha_{1,k}, \alpha_{2,k}$.}
	
	\For{$k=0,1,2\dots$}{
		\textbf{if} actor is leader, \textbf{then} update actor and critic in ${\tt ALG}$ with
		\vspace{-3mm}
		\begin{align*}
			\theta_{k+1} & = \theta_k + \alpha_{1,k} (\nabla_\theta J(\theta_k, w_{k,0})-(\nabla_{w\theta}^\top L \circ (\nabla_w^2L + \lambda I)^{-1} \circ \nabla_wJ)(\theta_k, w_{k,0}) ) \\
			w_{k,l+1} & = w_{k,l} - \alpha_{2,k} \nabla_w L(\theta_k, w_{k,l}),\quad l\in[0,m-1] \\
			w_{k+1,0} &= w_{k,m}
		\end{align*}
		\vspace{-7mm}
		
		\textbf{if} critic is leader, \textbf{then} update actor and critic in ${\tt ALG}$ with
		\vspace{-3mm}
		\begin{align*}
			w_{k+1} & = w_k - \alpha_{1,k} (\nabla_w L(\theta_{k,0}, w_k)-(\nabla_{\theta w}^\top J\circ (\nabla_\theta^2J+\lambda I)^{-1}\circ\nabla_\theta L)(\theta_{k,0}, w_k))\\
			\theta_{k,l+1} & = \theta_{k,l} + \alpha_{2,k} \nabla_\theta J(\theta_{k,l}, w_k), \quad l\in[0,m-1] \\
			\theta_{k+1,0}&=\theta_{k,m}
		\end{align*}
		\vspace{-5mm}
	}
	\caption{Stackelberg Actor-Critic Framework
		with Unrolling Follower Update and Regularization
	}
	\label{alg:framework_2}
\end{algorithm}

\end{document}